\documentclass[runningheads]{llncs}

\usepackage{eccv}

\usepackage{eccvabbrv}

\usepackage{graphicx}
\usepackage{booktabs}

\usepackage{algorithm}
\usepackage{tcolorbox}
\usepackage{xcolor}
\usepackage{algpseudocode}
\usepackage{booktabs}
\usepackage{multirow}
\usepackage{graphicx}
\usepackage{makecell}
\usepackage[numbers]{natbib}
\setcitestyle{square}

\usepackage[accsupp]{axessibility}  
\usepackage{etoolbox}
\AtBeginEnvironment{table}{\small}

\usepackage{hyperref}

\usepackage{orcidlink}

\begin{document}

\title{RankT2I: A Submodular Framework for Discovering Interpretable and Diverse Semantics in Text-to-Image Models} 

\titlerunning{RankT2I}

\author{Ritika Allada\inst{1}\orcidlink{0009-0003-5740-3011} \and
Pinar Yanardag\inst{1}\orcidlink{0009-0003-3452-7417}}

\authorrunning{R. Allada and P. Yanardag}
\institute{Virginia Tech, Blacksburg, VA, USA 
\\
\email{\{ritika88, pinary\}@vt.edu}}

\maketitle

\begin{abstract}
Recent advances in text-to-image (T2I) models have revolutionized the field of image generation and editing. However, identifying semantics that a T2I model can successfully edit in an image continues to be a challenging task. Most existing approaches require users to manually specify semantics to modify a particular image, a time-consuming process that often involves extensive trial and error. In this paper, we present RankT2I, a novel, training-free, and model-agnostic framework that automates the discovery of editable semantics in diffusion and FLUX-based models. Given a visual domain, we first utilize a multimodal vision-language model to gather a broad set of candidate semantics. We then frame semantic discovery as a set selection problem and use a submodular objective to identify semantics that are relevant, editable, and diverse. Our method helps users efficiently identify a wide range of semantics for text-to-image editing models across several domains while outperforming existing methods.

 
\end{abstract}

\section{Introduction}
\label{sec:intro}
Text-to-image (T2I) models \cite{rombach2022high, podell2024sdxl, zhang2023text, nichol2021glide, ramesh2022hierarchical, saharia2022photorealistic, xu2023versatile} have rapidly advanced the field of image generation and editing. In practice, these models are used to create stylistic changes \cite{han2025stylebooth, mazuz2025consistyle, chen2024artadapter} or add or remove fine-grained features in an image \cite{parihar2024precisecontrol, zhao2024ultraedit, baumann2025continuous}. Some applications of generative image editing include improving the lighting of photos \cite{butt2026lumictrl}, personalizing digital images \cite{dunlop2026personalized}, assisting users with virtual try-ons in the fashion industry \cite{li2026dit, choi2024improving}, and more.

\begin{figure}[h] 
  \centering
  \includegraphics[width=\linewidth]{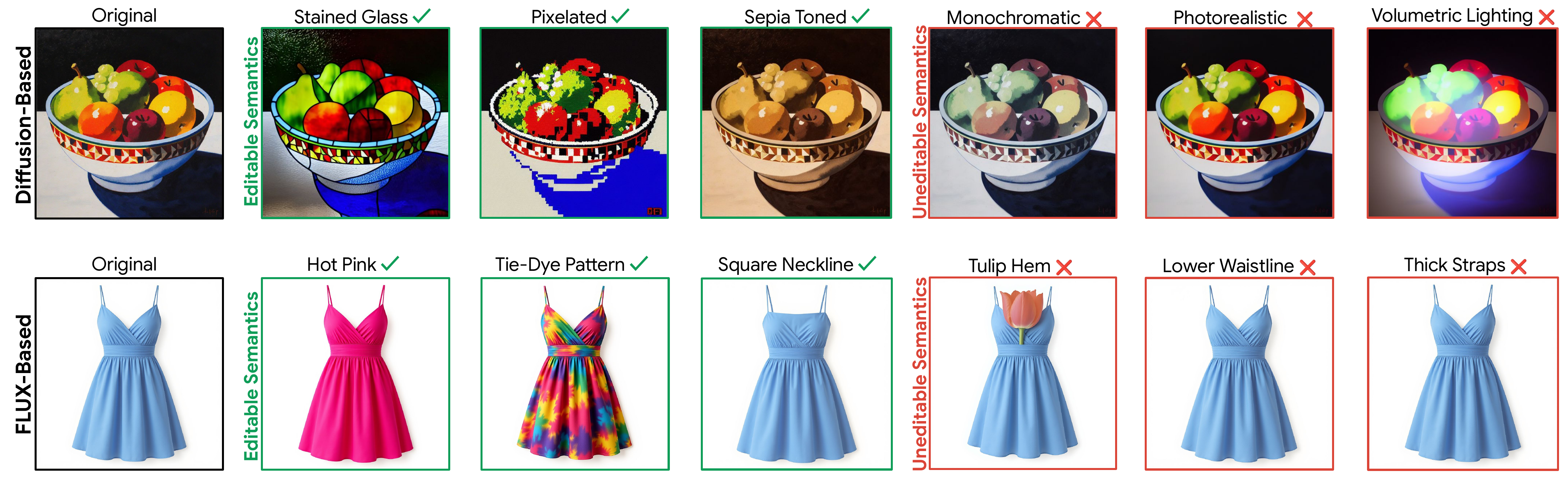}
  \caption{\textbf{Motivating Examples.} T2I editing methods like InstructPix2Pix \cite{brooks2023instructpix2pix} and FLUX.1 Kontext \cite{labs2025flux} can apply some semantics, but fail on others. This underscores the need for an automated approach to uncover the full range of editable semantics in generative models.}
  \label{fig:motivationfig}
\end{figure}

Recent works have focused on developing methods to improve the controllability of fine-grained image editing, such as manipulating specific latent features or using text prompts to guide the editing process \cite{dong2023prompt, orgad2023editing, kawar2023imagic, zhang2023sine, wu2025latentps, wu2023latent, jia2025designedit, ye2024progressive}. Although the controllability of T2I models has improved over the years \cite{li2019controllable, lin2024ctrl, meral2024conform, zhang2023adding}, discovering \textit{editable semantics} remains an ongoing challenge \cite{chen2026failureatlas, venkatesh2024ravel}. For example, as seen in the edited images with green borders in Figure \ref{fig:motivationfig}, some prompts (e.g., ``hot pink'' and ``tie-dye pattern'' in the dress domain) can be successfully edited by the underlying T2I model (e.g., InstructPix2Pix \cite{brooks2023instructpix2pix}, FLUX.1 Kontext \cite{labs2025flux}, etc.). However, other prompts cannot be edited by the model (e.g., ``tulip hem,'' ``lower waistline,'' etc. in the dress domain), as indicated by the edited images with red borders in Figure \ref{fig:motivationfig}. Thus, determining which aspects of an image are editable is a time-consuming process, particularly when it comes to crafting effective prompts, setting guidance scales, choosing suitable images, and tuning other model-specific parameters.

Finding a large set of editable semantics that are tailored to a specific domain is also an ongoing effort in T2I editing \cite{liu2025medebench}. Multimodal vision-language models can make this process easier by automating the generation \cite{qu2023layoutllm} of relevant semantics for a domain based on the user's prompt. However, using language models comes with its own set of limitations \cite{campbell2024understanding, zhang2025critic, izadi2026visual, arora2023have}, such as hallucinations or dependence on the original prompt \cite{marvin2023prompt, liu2023pre}, which can potentially influence the outputs of the downstream T2I model. Furthermore, it is also difficult for users to understand the full range of capabilities (e.g., creative image generation \cite{venkatesh2026crea, han2025enhancing, yang2026vibe}) offered by T2I editing models firsthand. Previous works have focused on improving the diversity of image generation models (e.g., GANs, diffusion, etc.) to reduce the production of similar outputs \cite{choi2020stargan, sonmezer2025loraverse, xia2021tedigan, zhang2022divergan}, while relatively few have explored the same problem in the context of image editing \cite{sivuk2025diverse, Simsar_2023_WACV}. Motivated by these challenges, we propose RankT2I as a potential solution to this problem. Our key contributions are as follows:

\begin{itemize}
\item We introduce RankT2I, a training-free, model-agnostic framework for automatically discovering and ranking editable semantics in T2I models. This allows users to identify and apply meaningful edits to their image using the model of their choice without requiring significant expertise.

\item We formulate semantic edit discovery as a set selection problem and propose a submodular objective that jointly optimizes semantic relevance, model editability, and visual diversity by leveraging text prompts with unedited and edited images.

\item We apply our approach across various domains (e.g., fashion, landscapes, artwork, and faces) and demonstrate through quantitative metrics that RankT2I consistently outperforms existing diffusion- and FLUX-based semantic discovery methods by finding more diverse and interpretable image edits in significantly less time.
\end{itemize}

\section{Related Work}

\subsection{Text-to-Image (T2I) Editing Models}
Text-to-image editing models have seen substantial progress in recent years, with diffusion- and FLUX-based models emerging as two dominant approaches. Diffusion-based models work by adding Gaussian noise over several time steps and then training a neural network to recreate the original image by reversing this process \cite{rombach2022high, song2020denoising, ho2020denoising}. Several diffusion-based T2I models have improved image editing by using implicit masking \cite{brack2024ledits++}, correcting edits by shifting the denoising process \cite{deutch2024turboedit}, using a stochastic differential equation to iteratively remove the noise from the generated image  \cite{meng2021sdedit}, training a T2I model based on synthetic images and text prompts \cite{brooks2023instructpix2pix}, using null-text optimization \cite{mokady2023null}, and incorporating conditioning layouts in T2I models \cite{zhang2023adding}.
 
Compared to diffusion models, rectified flow models can also be used to generate images by applying an ordinary differential equation (ODE) to connect two distributions \cite{lipman2022flow, liu2022flow, esser2024scaling}. A recent example of this is FLUX.1 Kontext, which depends on a velocity prediction objective to generate images \cite{labs2025flux}. Other examples of T2I rectified-flow-based models directly find mappings between original and edit text prompt distributions \cite{kulikov2025flowedit}, are used for inversion tasks \citep{rout2024semantic}, and apply both global and localized editing methods within FLUX \cite{dalva2025fluxspace}. Despite these advances, semantic discovery remains challenging, as some text prompts do not translate into editable visual changes.

\subsection{Finding Interpretable Directions in T2I Models}
Identifying interpretable directions for image editing is a difficult task and was first explored in generative adversarial networks (GANs) \cite{harkonen2020ganspace, yang2021discovering, hu2022unsupervised, yuksel2021latentclr, goetschalckx2019ganalyze, schwettmann2021toward}. Recently, a few works have explored applying semantic discovery to diffusion and FLUX-based T2I editing models. NoiseCLR \cite{dalva2024noiseclr} aims to find editable directions in diffusion models by training a contrastive learning objective. However, interpreting the directions identified by NoiseCLR is challenging because the resulting edits depend on the editing scale and the chosen denoising timesteps. Determining the appropriate scale and timestep settings for a given attribute can be difficult, particularly for users with limited T2I editing experience. Additionally, SliderSpace \cite{gandikota2025sliderspace} applies principal component analysis (PCA) to a set of generated images, extracts their CLIP features, and trains a slider for each principal component to identify interpretable directions. However, this process is inefficient, and the discovered semantics require labeling post-discovery. The quality of the resulting sliders also depends on the number of images generated for the target domain. For example, larger samples can lead to more robust directions but consequently slow down the semantic discovery process. Furthermore, selecting an appropriate editing scale is challenging, as large-scale values can potentially alter the identity of the original subject or distort the image.

Additionally, \cite{zhang2023unsupervised} finds interpretable directions in diffusion models by extending the framework proposed in \cite{voynov2020unsupervised} for GANs, but requires training. Furthermore, \cite{haas2024discovering} uses PCA and manipulates vectors in an h-space to identify semantic directions in diffusion models; however, this process relies on classifiers to label the edits, which may limit the directions that are discovered as each direction corresponds to a classifier-defined label. Compared to these existing methods, RankT2I is training-free and discovers several diverse, interpretable semantics across multiple domains for both diffusion and FLUX-based models.

\section{Methodology}

\begin{figure} 
  \centering
  \includegraphics[width=\linewidth]{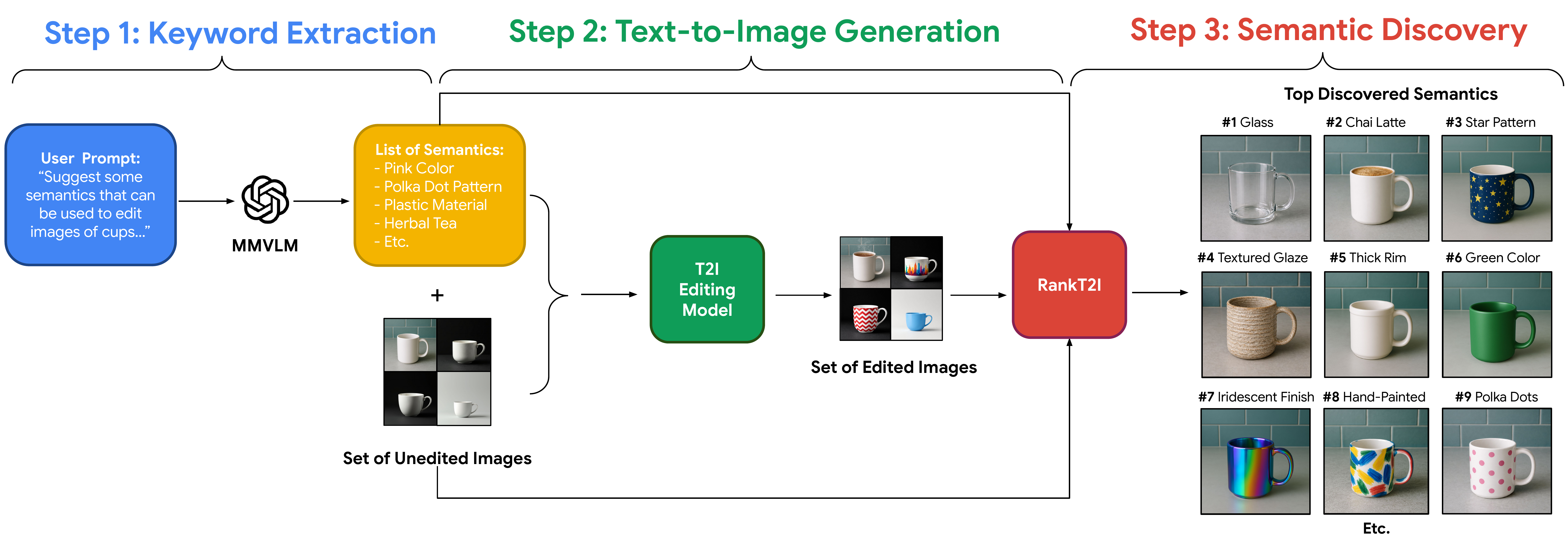}
  \caption{\textbf{An Overview of RankT2I.} First, we prompt a multimodal vision-language model (MMVLM) \cite{singh2025openai} to obtain a list of potentially editable semantics for a domain. Then, we feed these semantics and a set of unedited images into a T2I model \cite{brack2024ledits++, brooks2023instructpix2pix, labs2025flux}, to get a set of edited images. Lastly, we input the attributes list, unedited images, and edited images into our submodular framework, ``RankT2I,'' which suggests relevant, editable, and diverse semantics.}
  \label{fig:methoddiagram}
\end{figure}

To identify a set of relevant, editable, and diverse semantics for image editing, we follow a three-step process: (1) generate a list of candidate semantics, (2) create a set of edited images, and (3) use a submodular objective function to discover interpretable directions. Figure \ref{fig:methoddiagram} illustrates this process.

\subsection{Collection of Semantics}
Given a user prompt tailored to a specific domain, we use a multimodal vision-language model~\cite{singh2025openai} to generate a diverse set of candidate semantics for image editing. This produces keywords and phrases capturing a wide range of attributes, styles, and objects without manual specification. We denote this set of candidate semantics as \( \mathcal{T} = \{t_1, t_2, \dots, t_M\} \). To evaluate each semantic \( t_i \), we apply it to \( N \) original (unedited) images sampled from the domain.

Using a pre-trained T2I diffusion or FLUX-based model~\cite{brack2024ledits++, brooks2023instructpix2pix, labs2025flux}, we generate a set of edited images $E_i$ (where each edited image corresponds to a semantic \( t_i \in \mathcal{T} \) and where \(n_i\) is the total number of edited images generated for $t_i$):

\[E_i = \{x_{i,1}^{\text{edited}}, x_{i,2}^{\text{edited}}, \dots, x_{i,n_i}^{\text{edited}}\}\]

To determine how well a semantic can be edited within the target domain, we use the results from all of the edited images associated with that semantic.

\subsection{RankT2I}  

Given the candidate semantic set \( \mathcal{T} = \{t_1, t_2, \dots, t_M\} \), our goal is to select a subset $S$ that is representative and diverse. Presenting all candidates to the user is impractical when \( M \) is large (e.g., 500–1000), and many semantics may be redundant or correspond to minor visual variations. We therefore formulate semantic discovery as a subset selection problem: selecting \( S \subseteq \mathcal{T} \) with size \( K \ll M \) that prioritizes semantics that are relevant to the domain, reliably editable, and visually diverse. The resulting subset provides a concise summary of the editing space, offering users representative and high-quality edit options without overwhelming them with redundant suggestions. The subset selection algorithm itself operates entirely over $\mathcal{T}$, where each candidate corresponds to one semantic attribute.

We propose a submodular objective function that selects a \emph{relevant}, \emph{editable}, and \emph{diverse} subset (\( S \subseteq \mathcal{T} \)) of semantics for image editing, which is defined as follows:

\begin{equation}
F(S)=
\sum_{t_i \in S}\Bigl[
    \underbrace{\alpha F_{r_i}}_{\text{(Relevance)}}+
    \underbrace{\beta F_{e_i}}_{\text{(Editability)}}
\Bigr]+
\underbrace{\lambda \sum_{t_j \in \mathcal{T}} \max_{t_i \in S} s(t_i,t_j)}_{\text{(Diversity via coverage)}}.
\label{eqn:submodular_obj}
\end{equation}

In Equation \ref{eqn:submodular_obj}, $\alpha$, $\beta$, and $\lambda$ are weighting coefficients for relevance, editability, and diversity, respectively. Specifically, $t_i$ represents the candidate semantic that is currently being evaluated. On the other hand, $t_j$ represents a semantic attribute in the candidate set $\mathcal{T}$ and is used when measuring semantic coverage and diversity. The function $s(t_i,t_j)$ is the cosine similarity between the CLIP text embeddings of semantics $t_i$ and $t_j$, which is used to measure the coverage of various semantics. The key components of this submodular objective function are described below.
    
\textbf{Relevance ($F_{r_i}$):}  For a particular semantic $t_i$, we compute the CLIP similarity between every edited image generated using $t_i$ and the corresponding semantic $t_i$. The relevance score is the average over all edited images (where $j$ indexes the images):

\begin{equation}
F_{r_i}
=
\frac{1}{n_i}
\sum_{j=1}^{n_i}
\mathrm{CLIP}\!\left(x^{\mathrm{edited}}_{i,j},\, t_i\right),
\end{equation}

This measures how well the edits generated for semantic attribute $t_i$ align with the corresponding semantic description, averaged over all available edited samples associated with that semantic.

\textbf{Editability ($F_{e_i}$):} The editability score measures how well an edit is added to the original image while preserving the identity of the subject in the original image:

\begin{equation}
F_{e_i}
=
\frac{1}{n_i}
\sum_{j=1}^{n_i}
\left[
\mathrm{CLIP}\!\left(x^{\mathrm{edited}}_{i,j},\, t_i\right)
-
\mathrm{CLIP}\!\left(x^{\mathrm{orig}}_{j},\, t_i\right)
\right]
\mathrm{CLIP}\!\left(x^{\mathrm{edited}}_{i,j},\, x^{\mathrm{orig}}_{j}\right).
 \label{eqn:editability_score}
\end{equation}

The first term measures the increase in CLIP similarity to the semantic attribute $t_i$ introduced by the edit (e.g., how much more the edited image $x_{i,j}^{\mathrm{edited}}$ aligns with $t_i$ compared to the original image $x_j^{\mathrm{orig}}$). This gain is multiplied by $\mathrm{CLIP}(x_{i,j}^{\mathrm{edited}}, x_j^{\mathrm{orig}})$, the CLIP-based image-to-image similarity between the edited and original images, to encourage edits that preserve the visual content of the source image. Averaging this quantity over all available edited samples yields the editability score $F_{e_i}$ for semantic attribute $t_i$. A high $F_{e_i}$ therefore indicates that the semantic can be introduced consistently across images while maintaining strong visual similarity to the corresponding originals.

\begin{algorithm}
\caption{RankT2I}
\label{alg:greedy}
\begin{algorithmic}[1]
\State \textbf{Input:} Candidate semantics \( \mathcal{T}=\{t_1,\dots,t_M\} \), weights \(\alpha,\beta,\lambda\), budget \(K\)
\State \textbf{Output:} Ranked subset \(S \subseteq \mathcal{T}\), with \(|S|=K\)

\State Initialize \(S \gets \emptyset\)
\State Initialize \(m_j \gets 0,\ \forall t_j \in \mathcal{T}\) \Comment{Current coverage of each semantic}
\State Compute \(\Phi(t_i) \gets \alpha F_{r_i} + \beta F_{e_i},\ \forall t_i \in \mathcal{T}\)

\For{\(\ell = 1\) to \(K\)}
    \State \(t^* \gets \text{None},\ \Delta^* \gets -\infty\)
    \ForAll{\(t_i \in \mathcal{T} \setminus S\)}
        \State \(\Delta_{\mathrm{div}}(t_i \mid S) \gets
        \sum_{t_j \in \mathcal{T}}
        \left[
        \max\{s(t_i,t_j), m_j\} - m_j
        \right]\)
        \State \(\Delta F(t_i \mid S) \gets \Phi(t_i) + \lambda \Delta_{\mathrm{div}}(t_i \mid S)\)
        \If{\(\Delta F(t_i \mid S) > \Delta^*\)}
            \State \(\Delta^* \gets \Delta F(t_i \mid S)\)
            \State \(t^* \gets t_i\)
        \EndIf
    \EndFor
    \State \(S \gets S \cup \{t^*\}\)
    \State \(m_j \gets \max\{m_j, s(t^*,t_j)\},\ \forall t_j \in \mathcal{T}\)
\EndFor

\State \Return \(S\)
\end{algorithmic}
\end{algorithm}
\textbf{Diversity:} The second summation term is a diversity reward, defined as a facility-location function \cite{lin2011class} that encourages the selected set \( S \) to cover the entire semantic space \( \mathcal{T} \); \( \lambda \) controls its contribution. To obtain the ranked list of semantic edits, we maximize $F(S)$ under a budget (e.g., select the top $K$ edits to present). To optimize this function, we adopt a greedy strategy that is optimal up to a known constant factor $(1 - 1/e) \approx 63\%$ \cite{sviridenko2004note, nemhauser1978analysis}. The greedy algorithm starts with $S=\emptyset$ and iteratively adds the candidate $t_i \notin S$ that yields the largest marginal gain (e.g., $\Delta F = F(S \cup {t_i}) - F(S)$). At each iteration, only semantic attributes that have not yet been selected are considered; we continue this process until $|S| = K$. Because $F$ is submodular, each addition has a \textit{diminishing returns} effect: once a certain semantic niche (e.g., \textit{edits related to sleeve types in the dress domain}) is filled by one concept, other similar concepts will have lower marginal gains. Note that since the editability score can be negative when an edit reduces alignment with the target semantic, we clip each candidate's combined score to be non-negative. This ensures our function is monotone in addition to submodular, so the greedy procedure retains its approximation guarantee.

To perform the selection, we maintain for every semantic attribute $t_j \in \mathcal{T}$ a running coverage value:

\begin{equation}
m_j=\max_{t_{i'}\in S}s(t_{i'},t_j),
\label{eq:mj_def}
\end{equation}

which is initialized to $m_j=0$ when $S=\emptyset$. At each iteration, we evaluate the marginal gain of each remaining semantic attribute $t_i \notin S$ as

\begin{equation}
\Delta F(t_i \mid S) = \Phi(t_i) + \lambda \sum_{t_j\in\mathcal{T}} \left[
\max\{s(t_i,t_j),\,m_j\} - m_j \right]
\label{eq:full_rewrite}
\end{equation}

In Equation \ref{eq:full_rewrite}, $\Phi(t_i)=\alpha F_{r_i}+\beta F_{e_i}$ denotes the combined relevance and editability score associated with semantic attribute $t_i$. This computation is efficient since for each semantic attribute $t_j$, the diversity contribution is:

\[
\lambda\left(\max\{s(t_i,t_j),\,m_j\}-m_j\right),
\]

which is positive only when selecting $t_i$ improves the current coverage of $t_j$. After selecting the semantic attribute with the largest marginal gain, we incrementally update the coverage values as:

\[
m_j \leftarrow \max\{m_j,\; s(t_i,t_j)\}, \qquad \forall\, t_j\in\mathcal{T}.
\]

We repeat this process until the selected set satisfies $|S| = K$ and finally output the ranked list of semantics to the user. The pseudocode is outlined in Algorithm \ref{alg:greedy}.

\begin{figure*}[!t]
  \centering
  \includegraphics[width=\textwidth]{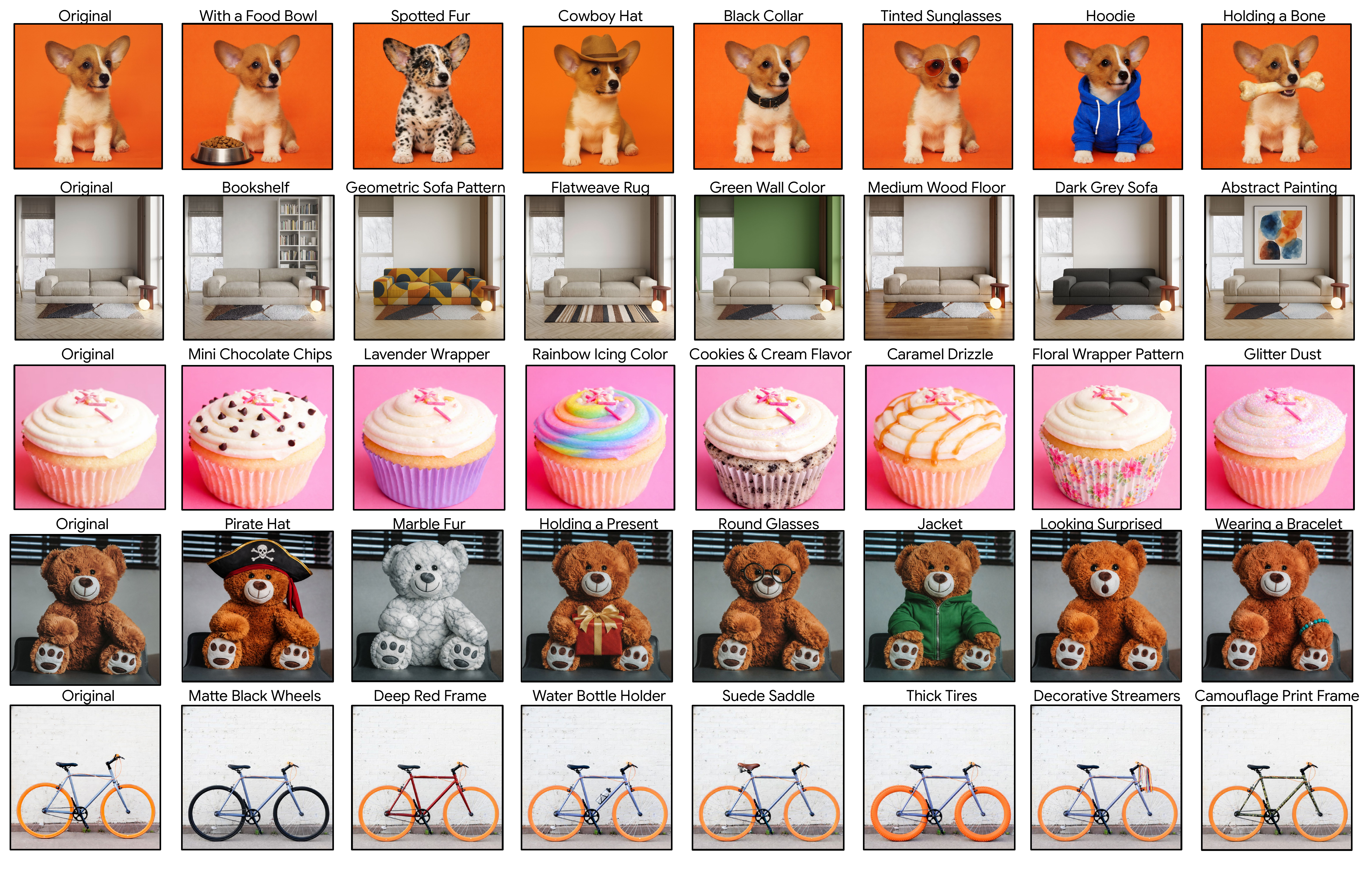}
  \caption{\textbf{Semantic Discovery for Real-Image Editing Using a Closed-Source Model.} RankT2I finds both diverse and interpretable semantics in real images in many domains. GPT-Image-1.5 (see Footnote \ref{fn:gptimg15}) was the base model that was used to generate these edits. The original images are from Unsplash (see Footnote \ref{fn:unsplash}).}
  \label{fig:top10semantics3}
\end{figure*}

\section{Experiments}
We demonstrate RankT2I's effectiveness through three core tasks: (1) uncovering a broad set of diverse, relevant, and editable semantics across multiple domains, (2) qualitatively and quantitatively assessing the diversity and interpretability of the discovered attributes against existing semantic discovery and attribute retrieval methods, and (3) performing ablation studies to reveal how hyperparameters and model choices impact the final rankings. In the supplementary material, we include experiments demonstrating the generalizability of these semantics across images and additional experimental details (e.g., prompts).

\begin{figure*}[!t]
  \centering
   \includegraphics[width=\textwidth]{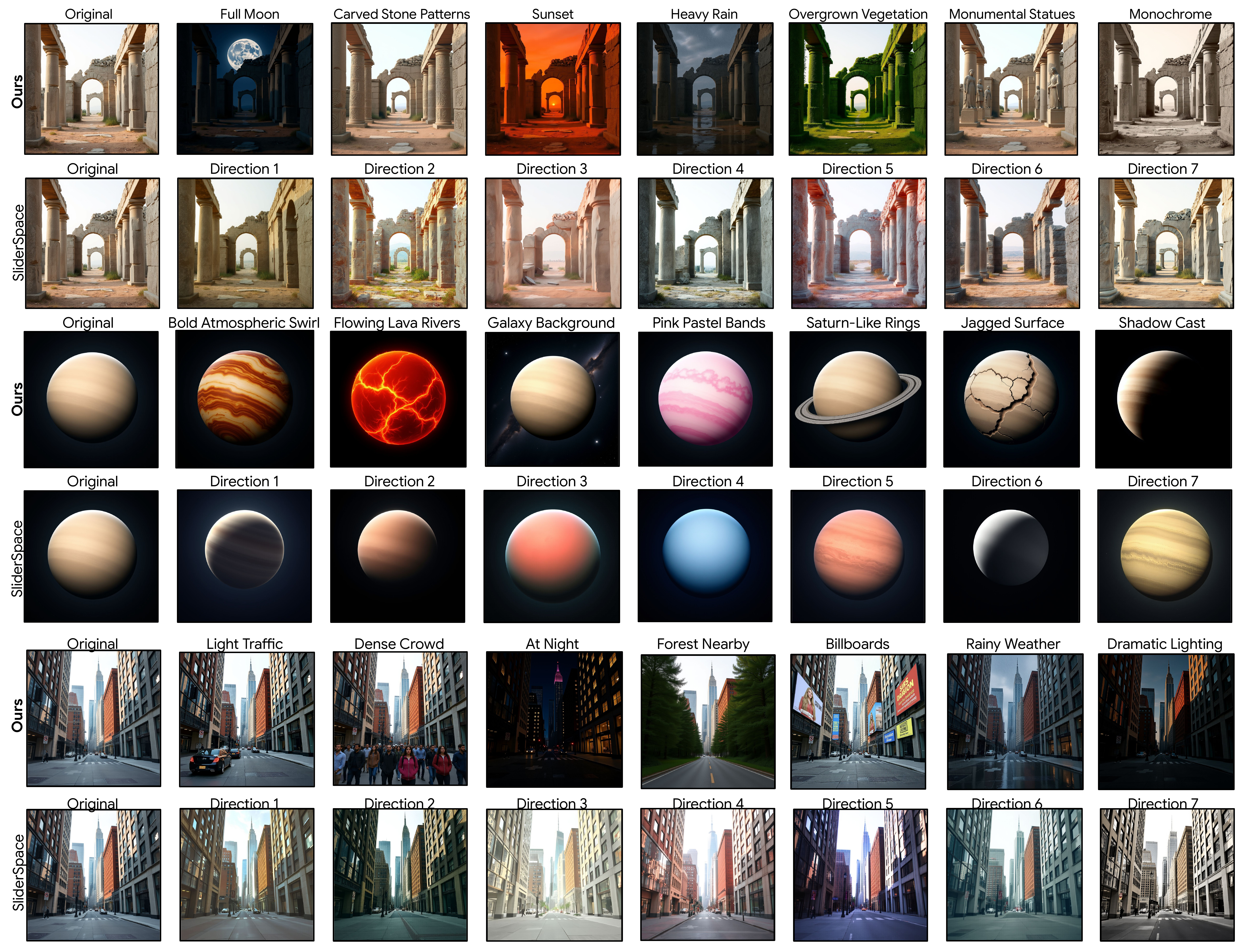}
  \caption{\textbf{Top Discovered Semantics in FLUX-Based Models.} RankT2I successfully identifies several semantics in FLUX-based T2I models compared to SliderSpace across various domains, including ancient ruins, planets, and cityscapes. }
  \label{fig:top10semantics2}
\end{figure*}

\noindent \textbf{Experimental Setup.} All experiments were performed using an NVIDIA L40 GPU, with the exception of the generated SliderSpace FLUX images in Figure \ref{fig:top10semantics2} and the FLUX-based time comparisons in Table \ref{tab:total_time}. In these cases, an NVIDIA H100 SXM5 GPU was used. Additionally, for RankT2I, the diffusion-based edited images were created using Ledits++ \cite{brack2024ledits++}, and InstructPix2Pix \cite{brooks2023instructpix2pix}, while FLUX.1 Kontext \cite{labs2025flux} was used to generate the FLUX-based edited images. For Figure \ref{fig:top10semantics3}, we showcase RankT2I's ability to successfully find editable semantics in real images using a closed-source model. In this figure, GPT-Image-1.5 \footnote{\label{fn:gptimg15}https://developers.openai.com/api/docs/models/gpt-image-1.5} was used as the editing model, and the free, original images were obtained from Unsplash \footnote{\label{fn:unsplash}https://unsplash.com/}. Unless otherwise noted, the rest of the original images in the figures were generated using GPT-Image-1.5 (see Footnote \ref{fn:gptimg15}), SDXL \cite{podell2024sdxl}, and FLUX Schnell \footnote{https://huggingface.co/black-forest-labs/FLUX.1-schnell} \cite{Labs_2024}. 

As for the images used in RankT2I, we used $N$=5 unedited images per domain in Figures \ref{fig:top10semantics2} and \ref{fig:top10semantics}; as for Figure \ref{fig:top10semantics3}, we generated edits for a single image ($N$=1) to show how RankT2I can be used to find semantics specific to a particular image rather than an entire domain. 

\subsection{Qualitative Experiments}

\begin{figure*}[!t]
  \centering
  \includegraphics[width=\textwidth]{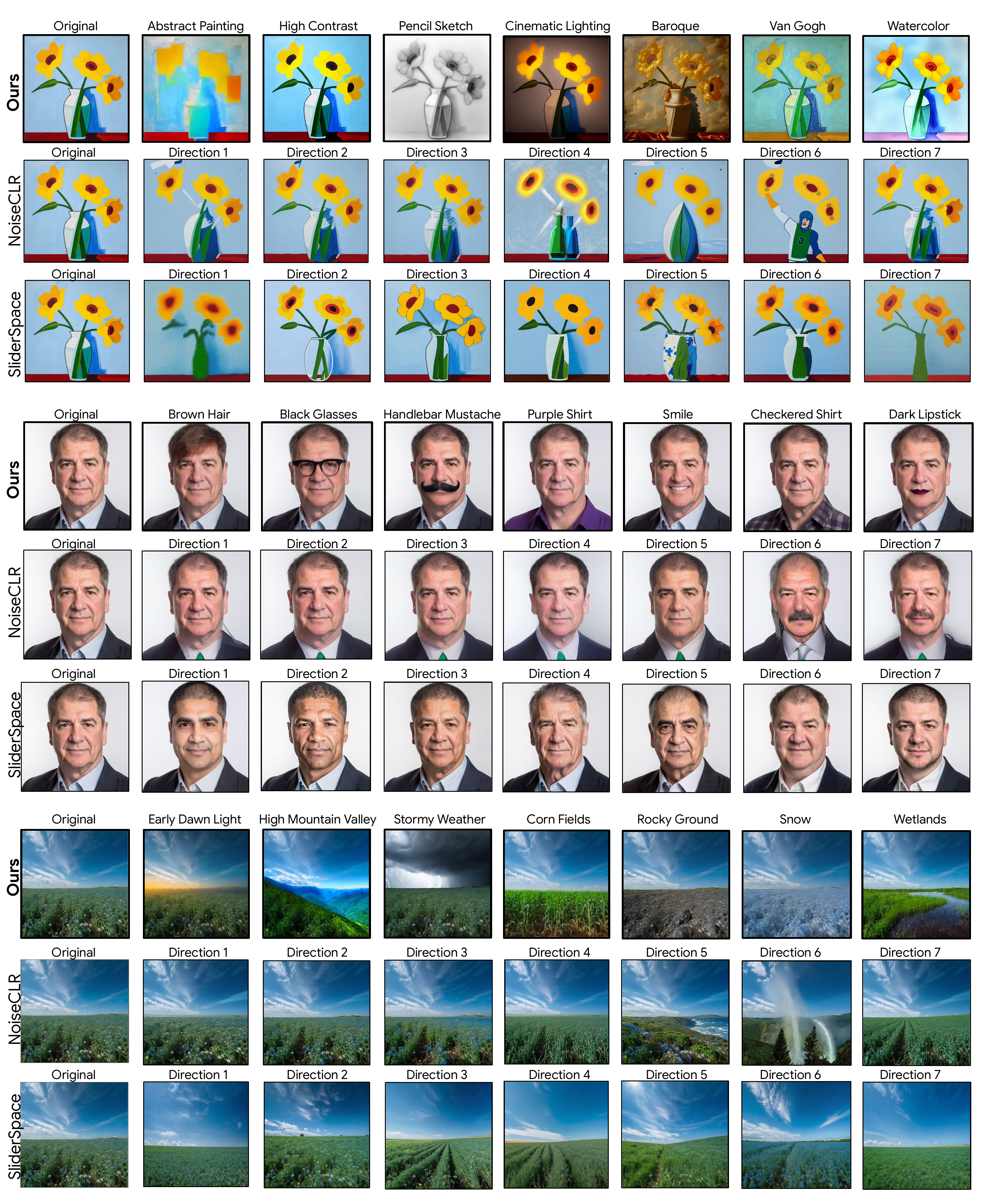}
  \caption{\textbf{Top Discovered Semantics in Diffusion-Based Models.} Compared to existing diffusion-based semantic discovery methods, like NoiseCLR and SliderSpace, RankT2I successfully identifies a wide range of diverse and interpretable semantics. }
  \label{fig:top10semantics}
\end{figure*}

Figures \ref{fig:top10semantics2} and \ref{fig:top10semantics} illustrate that RankT2I consistently identifies a diverse, relevant, and highly editable set of semantics across a wide range of domains, including landscapes, art, faces, etc., compared to existing diffusion-based and FLUX-based semantic discovery methods. Specifically, we compared RankT2I with other semantic discovery methods that have publicly available code (NoiseCLR and SliderSpace). Since these methods do not rank their discovered directions, we used the first-K tokens or sliders to generate the top images. For example, in the face domain in diffusion models, RankT2I suggests a wide-variety of attributes such as hair color, shirt pattern, mustache type, etc., all of which can be directly edited with the selected T2I model (Ledits++  \cite{brack2024ledits++} in this case).  Unlike RankT2I, some directions appear visually similar across different tokens and sliders in NoiseCLR and SliderSpace. For example, in the landscapes domain, directions 3 and 4 from SliderSpace both appear to add rows to the field of grass, while in the face domain, directions 1–3 from NoiseCLR all appear to add a green tie to the man (see Figure  \ref{fig:top10semantics}). RankT2I also identifies diverse attributes in various domains in FLUX T2I models. For instance, in the cityscapes and ancient ruins domains, RankT2I discovers a range of edits such as different times of day (e.g., ``at night'' and ``sunset''), whereas the directions discovered by SliderSpace mainly appear to modify the color tone of the image in Figure \ref{fig:top10semantics2}. Additionally, Figure \ref{fig:dressgen} shows that our method discovers semantic directions that can be used to edit various images across a domain (and are not limited to a single image).  

\begin{figure*}[!t]
  \centering
  \includegraphics[width=\textwidth]{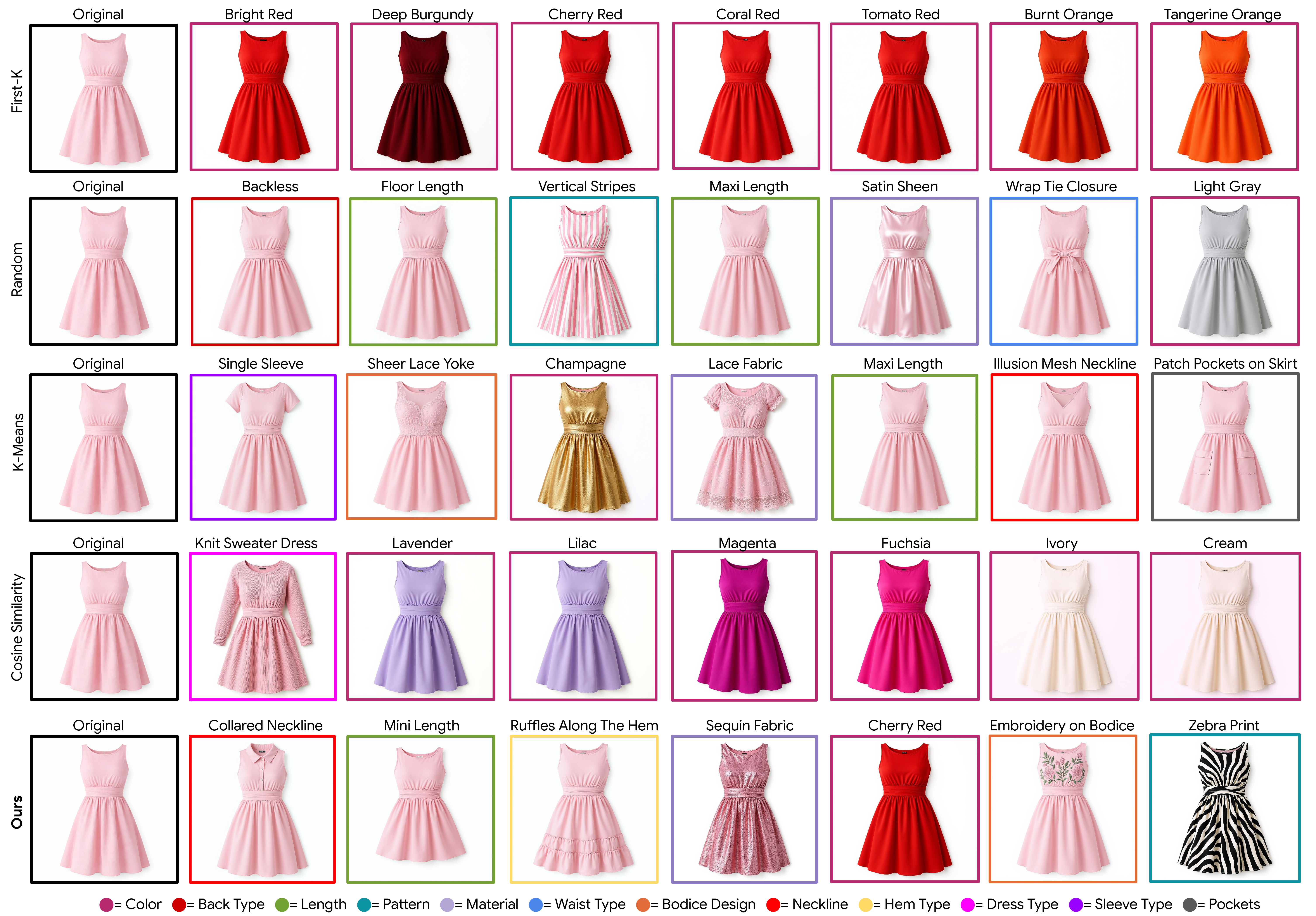}
  \caption{\textbf{Comparisons with Other Attribute Retrieval Methods.} RankT2I successfully identifies editable, diverse, and interpretable semantics compared to other attribute retrieval methods (e.g., first-K images, random selection, k-means, etc.).}
  \label{fig:top10semantics4}
\end{figure*}

\begin{figure}[h] 
  \centering
  \includegraphics[width=\linewidth]{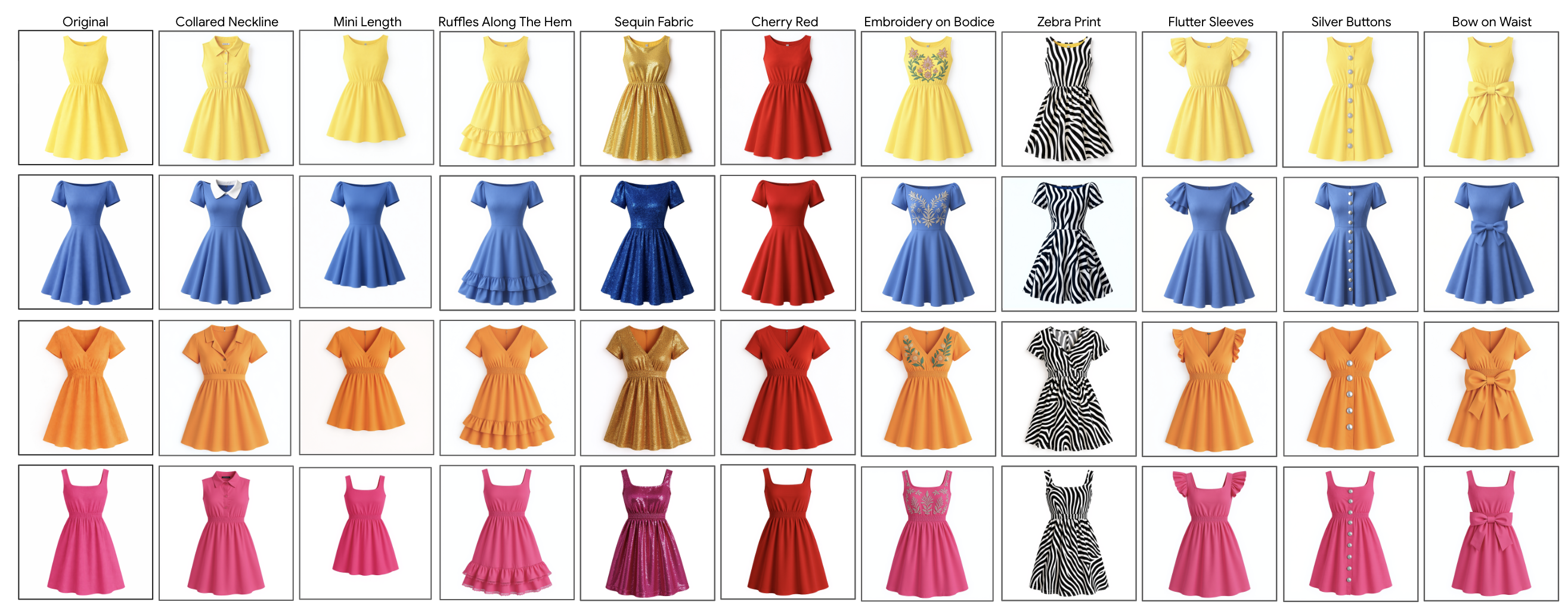}
  \caption{\textbf{Generalizability of Semantics.} The top semantics discovered by RankT2I can be applied to edit different images (Base Model: FLUX.1 Kontext).}
  \label{fig:dressgen}
\end{figure}

\subsection{Quantitative Experiments}
To quantitatively evaluate RankT2I’s ability to discover diverse and interpretable attributes for diffusion-based and FLUX-based image editing, we compared it with NoiseCLR and SliderSpace using several benchmarks. For a fair comparison, we generated 5,000 images per method (25 directions, 200 images each) for a total of 15,000 images in the diffusion-based comparisons and 10,000 images in the FLUX-based comparisons. We assessed the diversity of the discovered attributes using the truncated CLIP entropy (TCE) \cite{ibarrola2024measuring}, truncated inception entropy (TIE) \cite{ibarrola2024measuring}, and pairwise image similarity using image-to-image (I2I) CLIP \cite{hessel2021clipscore} scores. TCE measures the diversity of various semantics in CLIP's embedding space, TIE calculates the varied visual appearance through InceptionV3 \cite{szegedy2016rethinking}, and I2I finds the pairwise similarity between sets of edited images using CLIP. When interpreting TCE and TIE scores, higher values indicate images with higher diversity; for I2I, lower values indicate more visually diverse sets of images. 

Furthermore, we evaluated the interpretability of the discovered semantics using the CLIP-T \cite{radford2021learning}, TIFA \cite{hu2023tifa}, and VQAScore \cite{lin2024evaluating} metrics. CLIP-T finds the semantic alignment between an image and a text description through a contrastive objective. VQAScore measures text–image alignment by asking a VQA model a yes-or-no question about whether the image depicts the prompt, whereas TIFA measures alignment by generating multiple questions about specific objects, attributes, and relationships in the prompt and verifying each with a VQA model. We used the discovered semantics from the original prompts during the image generation process as the text descriptions when calculating the CLIP-T, TIFA, and VQAScore metrics for RankT2I. Since NoiseCLR and SliderSpace depend on labeling the semantics that describe the edited images after they are generated, we utilized GPT-5 \cite{singh2025openai} to generate captions describing the discovered semantics as the text prompt. As seen in Table \ref{tab:tab2}, RankT2I is able to find a balance between diverse and interpretable semantics, consistently outperforming NoiseCLR and SliderSpace across these metrics.

Besides semantic discovery methods, we compared RankT2I with attribute retrieval approaches that select attributes for image editing through different strategies. For example, when editing an image, users can select semantics in several ways: randomly sampling from a list, using cosine similarity between the domain prompt and attributes to identify the most relevant ones, selecting the first-K attributes from a prompts list, or clustering attributes with k-means \cite{MacQueen1967} and sampling one attribute from each cluster to encourage diversity. Our approach differs by examining the edited images produced by the target T2I model to uncover semantics that are specific to the model, rather than relying on predefined keywords. For both diffusion and FLUX-based models, we generated 5,000 images per attribute retrieval method (25 directions with 200 images each), totaling 25,000 images for the diffusion experiments and 25,000 images for the FLUX-based model experiments. Table \ref{tab:tab1} shows that RankT2I identifies both diverse and interpretable semantics across various T2I models compared to these attribute retrieval methods. Also, as seen in Figure \ref{fig:top10semantics4}, in the dress domain, selecting the first-K edited images or choosing attributes with the highest cosine similarity to a domain's text prompt often results in redundant attributes from a single broad category (e.g., colors). Random selection can also produce redundancy (e.g., dress lengths) and may suggest attributes that the base model cannot reliably edit (e.g., “backless”). Similarly, k-means clustering may include uneditable attributes (e.g., “maxi length” or “single sleeve”). In contrast, RankT2I finds semantics that are both editable and diverse (e.g., variations in necklines, hem types, bodice designs, etc. in the dress domain). 

Additionally, we compared the time it takes for our method to discover 100 semantics to NoiseCLR and SliderSpace in Table \ref{tab:total_time}. Regardless of the base model (e.g., diffusion, FLUX), RankT2I takes significantly less time to find semantics for image editing than existing methods. This disparity in the total time can primarily be attributed to the training-free nature of our method.

\begin{table}[t]
\centering
\footnotesize
\setlength{\tabcolsep}{3.5pt}
\renewcommand{\arraystretch}{1.05}
\begin{tabular}{llcccccc}
\toprule
\textbf{Model} & \textbf{Method}
& \textbf{$\downarrow$I2I}
& \textbf{$\uparrow$TCE}
& \textbf{$\uparrow$TIE}
& \textbf{$\uparrow$CLIP-T}
& \textbf{$\uparrow$TIFA}
& \textbf{$\uparrow$VQA} \\
\midrule
SD & NoiseCLR
& \underline{0.8439} & 13.8819 & \underline{34.2546}
& \underline{0.2378} & 0.3402 & \underline{0.5484} \\
SD & SliderSpace
& 0.8682 & \underline{16.5061} & 32.9884
& 0.2355 & \underline{0.3872} & 0.5236 \\
SD & \textbf{Ours}
& \textbf{0.7800} & \textbf{26.4701} & \textbf{41.1987}
& \textbf{0.2595} & \textbf{0.6078} & \textbf{0.8177} \\
\midrule
FLUX & SliderSpace
& \underline{0.9210} & \underline{12.5769} & \underline{35.2775}
& \underline{0.2542} & \underline{0.3462} & \underline{0.6737} \\
FLUX & \textbf{Ours}
& \textbf{0.8405} & \textbf{22.8102} & \textbf{40.7330}
& \textbf{0.2840} & \textbf{0.7159} & \textbf{0.9206} \\
\bottomrule
\end{tabular}
\caption{\textbf{Quantitative Comparisons with Semantic Discovery Methods.}
RankT2I identifies more diverse and interpretable attributes in both diffusion- and FLUX-based T2I models compared to existing methods.}
\label{tab:tab2}
\end{table}

\begin{table}[t]
\centering
\footnotesize
\setlength{\tabcolsep}{3.5pt}
\renewcommand{\arraystretch}{1.05}
\begin{tabular}{llcccccc}
\toprule
\textbf{Model} & \textbf{Method}
& \textbf{$\downarrow$I2I}
& \textbf{$\uparrow$TCE}
& \textbf{$\uparrow$TIE}
& \textbf{$\uparrow$CLIP-T}
& \textbf{$\uparrow$TIFA}
& \textbf{$\uparrow$VQA} \\
\midrule
SD & First-K
& 0.9039 & 12.5140 & 26.4074 & 0.2577 & 0.2290 & 0.7012 \\
SD & Random
& 0.8611 & 14.5577 & 32.4777 & 0.2595 & 0.3356 & 0.6972 \\
SD & K-Means
& \textbf{0.8433} & \underline{19.1433} & \underline{34.5400}
& 0.2635 & 0.4156 & \underline{0.7592} \\
SD & Cos. Sim.
& 0.9002 & 13.2874 & 27.5639 & \textbf{0.2764}
& \underline{0.4846} & 0.7459 \\
SD & \textbf{Ours}
& \underline{0.8436} & \textbf{19.6326} & \textbf{35.5105}
& \underline{0.2688} & \textbf{0.5021} & \textbf{0.8048} \\
\midrule
FLUX & First-K
& 0.8962 & 12.6479 & 30.7635 & 0.2631 & 0.4224 & 0.7028 \\
FLUX & Random
& 0.9360 & 4.6556 & 26.2837 & 0.2450 & 0.4292 & 0.6156 \\
FLUX & K-Means
& 0.8826 & \underline{18.1152} & 35.1952 & 0.2668 & 0.4216 & 0.7936 \\
FLUX & Cos. Sim.
& \textbf{0.8518} & 17.3142 & \underline{36.0747}
& \textbf{0.2920} & \underline{0.5840} & \underline{0.7962} \\
FLUX & \textbf{Ours}
& \underline{0.8566} & \textbf{20.8757} & \textbf{37.9963}
& \underline{0.2799} & \textbf{0.5995} & \textbf{0.8948} \\
\bottomrule
\end{tabular}
\caption{\textbf{Quantitative Comparisons with Attribute Retrieval Methods.}
RankT2I discovers more diverse and interpretable semantics than the majority of these attribute retrieval methods.}
\label{tab:tab1}
\end{table}

\begin{figure}[h] 
  \centering
  \includegraphics[width=\linewidth]{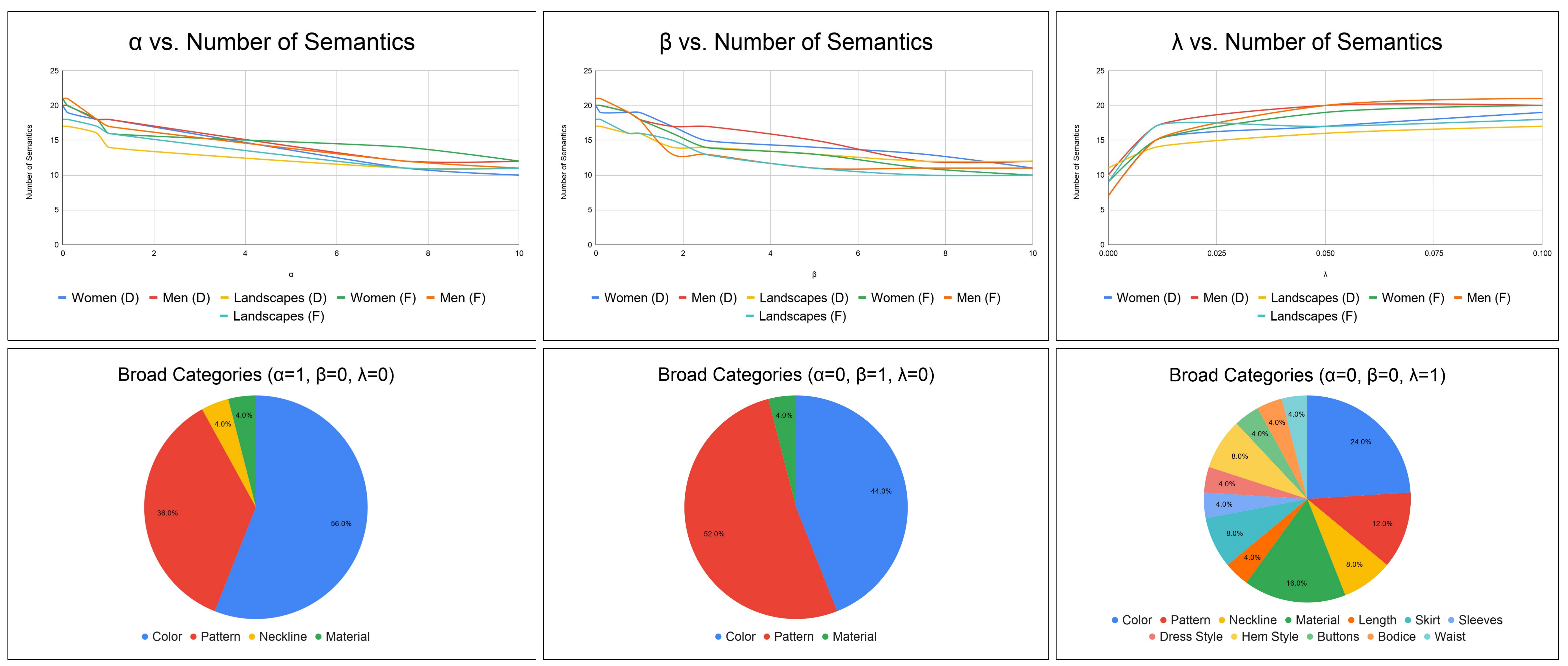}
  \caption{\textbf{Hyperparameter Ablation Studies.} These charts show how changing the hyperparameters ($\alpha$, $\beta$, and $\lambda$) affects the number of categories suggested by RankT2I. Note: (D) and (F) in the line graphs indicate images that were generated using diffusion- and FLUX-based T2I models.}
  \label{fig:ablations1}
\end{figure}

\subsection{Ablation Studies}
To analyze how hyperparameters influence RankT2I, we conducted ablation studies examining the effects of $\lambda$ (diversity), $\beta$ (editability), and $\alpha$ (relevance) on the number of categories represented in the top 25 ranked attributes across multiple domains and in both diffusion and FLUX-based models. The line graphs in Figure~\ref{fig:ablations1} show how the number of attribute categories changes as each hyperparameter varies. When analyzing the effect of a single hyperparameter, the remaining parameters are fixed to 1. As shown in the line graphs, increasing $\lambda$ leads to a higher number of represented categories, while increasing $\alpha$ or $\beta$ reduces the number of categories. The pie charts below the line graphs in Figure~\ref{fig:ablations1} further illustrate how the rankings change when optimizing for only one objective in the dress domain using FLUX.1 Kontext. Ranking by relevance or editability favors a narrow set of categories, whereas diversity-based ranking improves coverage but may select less editable attributes. Overall, these results demonstrate that a balanced combination of $\alpha$, $\beta$, and $\lambda$ is necessary for RankT2I to recommend attributes that are simultaneously meaningful, editable, and diverse.

\begin{table}[t]
\centering
\footnotesize
\setlength{\tabcolsep}{4pt}
\renewcommand{\arraystretch}{1.05}
\begin{tabular}{llccccc}
\toprule
\textbf{Model} & \textbf{Method}
& \textbf{Keyword}
& \textbf{Training}
& \textbf{Generation}
& \textbf{Ranking}
& \textbf{Total} \\
\midrule
SD & NoiseCLR
& 0
& 1114m 3s
& 3m 11s
& 0
& 1117m 14s \\
SD & SliderSpace
& 0
& 389m 33s
& 2m 15s
& 0
& 391m 48s \\
SD & \textbf{Ours}
& 24s
& 0
& 42m 21s
& 29s
& \textbf{43m 14s} \\
\midrule
FLUX & SliderSpace
& 0
& 287m 36s
& 3m 1s
& 0
& 290m 37s \\
FLUX & \textbf{Ours}
& 25s
& 0
& 112m 40s
& 1m 13s
& \textbf{114m 18s} \\
\bottomrule
\end{tabular}
\caption{\textbf{Semantic Discovery Time Comparisons.}
RankT2I discovers 100 semantics for text-to-image editing significantly faster than existing methods.}
\label{tab:total_time}
\end{table}

\begin{figure}[h] 
  \centering
  \includegraphics[width=1\linewidth]{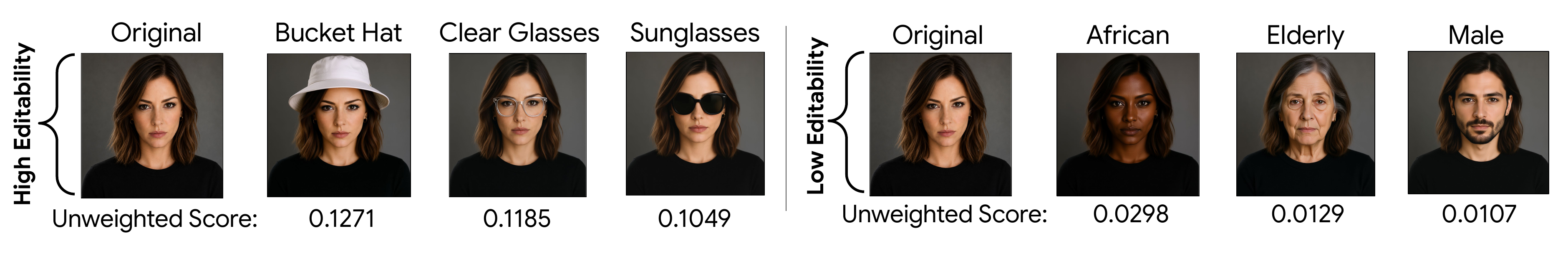}
  \caption{\textbf{Editability Score vs. Identity Preservation.} As shown above, semantics with editability scores below approximately 0.03 may alter the identity of the subject.}
  \label{fig:deepfakeexp}
\end{figure}

\section{Social Implications}

Like other T2I methods, RankT2I could potentially be misused for inappropriate content or deepfake generation. To help mitigate these risks, NSFW filtering during MMVLM-based keyword extraction or incorporating editing models that aim to preserve subject identity through disentangled modifications (such as Ledits++ with its implicit masking mechanism) can reduce the potential for generating harmful content. To further investigate the potential of RankT2I for mitigating deepfake generation, we conducted an experiment in the face-editing domain using FLUX.1 Kontext \cite{labs2025flux}. We generated 100 image edits and ranked them solely according to their editability scores. Our results suggest that, within this domain, semantics associated with editability scores below approximately 0.03 may be more likely to alter the identity of the original subject (see Figure \ref{fig:deepfakeexp}). However, this threshold likely depends on several factors (such as the T2I editing model, domain, and set of text prompts) and may vary for other T2I editing systems. Nevertheless, the editability score provides a promising quantitative measure for identifying semantics that pose a higher risk of identity modification, potentially helping practitioners establish safeguards.

\section{Conclusion}
We present RankT2I, a framework for discovering diverse, editable, and relevant semantics in T2I models. Specifically, we leverage a multimodal vision-language model to suggest a list of candidate semantics, generate edited images based on those semantics, and then apply a novel, submodular objective function to identify a set of semantics to present to users. Our approach can be applied to any T2I editing model (e.g., diffusion-based, FLUX-based, etc.) and is training-free, which significantly reduces the time needed to identify such semantics. Our experiments across multiple domains show that RankT2I uncovers a broader and more interpretable set of semantics than existing approaches, enabling applications such as stylistic image transformations as well as fine-grained object- and scene-level editing.

\section*{Acknowledgments}
We thank Yusuf Dalva and Enis Simsar for their support during the early stages of this research project. This research is supported by the National Science Foundation under Grant No.\ 2543524. 

\bibliographystyle{splncs04}
\bibliography{main}

@String(AAAI  = {AAAI})

@String(TOG   = {ACM Trans. Graph.})

@String(TOG   = {ACM TOG})

@article{zhou2017places,
  title={Places: A 10 million image database for scene recognition},
  author={Zhou, Bolei and Lapedriza, Agata and Khosla, Aditya and Oliva, Aude and Torralba, Antonio},
  journal={IEEE transactions on pattern analysis and machine intelligence},
  volume={40},
  number={6},
  pages={1452--1464},
  year={2017},
  publisher={IEEE}
}

@inproceedings{brack2024ledits++,
  title={Ledits++: Limitless image editing using text-to-image models},
  author={Brack, Manuel and Friedrich, Felix and Kornmeier, Katharia and Tsaban, Linoy and Schramowski, Patrick and Kersting, Kristian and Passos, Apolin{\'a}rio},
  booktitle={Proceedings of the IEEE/CVF conference on computer vision and pattern recognition},
  pages={8861--8870},
  year={2024}
}

@article{nichol2021glide,
  title={Glide: Towards photorealistic image generation and editing with text-guided diffusion models},
  author={Nichol, Alex and Dhariwal, Prafulla and Ramesh, Aditya and Shyam, Pranav and Mishkin, Pamela and McGrew, Bob and Sutskever, Ilya and Chen, Mark},
  journal={arXiv preprint arXiv:2112.10741},
  year={2021}
}

@article{labs2025flux,
  title={Flux. 1 kontext: Flow matching for in-context image generation and editing in latent space},
  author={Labs, Black Forest and Batifol, Stephen and Blattmann, Andreas and Boesel, Frederic and Consul, Saksham and Diagne, Cyril and Dockhorn, Tim and English, Jack and English, Zion and Esser, Patrick and others},
  journal={arXiv preprint arXiv:2506.15742},
  year={2025}
}

@inproceedings{dalva2025fluxspace,
  title={FluxSpace: Disentangled Semantic Editing in Rectified Flow Models},
  author={Dalva, Yusuf and Venkatesh, Kavana and Yanardag, Pinar},
  booktitle={Proceedings of the Computer Vision and Pattern Recognition Conference},
  pages={13083--13092},
  year={2025}
}

@article{rout2024semantic,
  title={Semantic image inversion and editing using rectified stochastic differential equations},
  author={Rout, Litu and Chen, Yujia and Ruiz, Nataniel and Caramanis, Constantine and Shakkottai, Sanjay and Chu, Wen-Sheng},
  journal={arXiv preprint arXiv:2410.10792},
  year={2024}
}

@inproceedings{kulikov2025flowedit,
  title={Flowedit: Inversion-free text-based editing using pre-trained flow models},
  author={Kulikov, Vladimir and Kleiner, Matan and Huberman-Spiegelglas, Inbar and Michaeli, Tomer},
  booktitle={Proceedings of the IEEE/CVF International Conference on Computer Vision},
  pages={19721--19730},
  year={2025}
}

@inproceedings{zhang2023adding,
  title={Adding conditional control to text-to-image diffusion models},
  author={Zhang, Lvmin and Rao, Anyi and Agrawala, Maneesh},
  booktitle={Proceedings of the IEEE/CVF international conference on computer vision},
  pages={3836--3847},
  year={2023}
}

@inproceedings{mokady2023null,
  title={Null-text inversion for editing real images using guided diffusion models},
  author={Mokady, Ron and Hertz, Amir and Aberman, Kfir and Pritch, Yael and Cohen-Or, Daniel},
  booktitle={Proceedings of the IEEE/CVF conference on computer vision and pattern recognition},
  pages={6038--6047},
  year={2023}
}

@inproceedings{brooks2023instructpix2pix,
  title={Instructpix2pix: Learning to follow image editing instructions},
  author={Brooks, Tim and Holynski, Aleksander and Efros, Alexei A},
  booktitle={Proceedings of the IEEE/CVF conference on computer vision and pattern recognition},
  pages={18392--18402},
  year={2023}
}

@article{meng2021sdedit,
  title={Sdedit: Guided image synthesis and editing with stochastic differential equations},
  author={Meng, Chenlin and He, Yutong and Song, Yang and Song, Jiaming and Wu, Jiajun and Zhu, Jun-Yan and Ermon, Stefano},
  journal={arXiv preprint arXiv:2108.01073},
  year={2021}
}

@inproceedings{deutch2024turboedit,
  title={Turboedit: Text-based image editing using few-step diffusion models},
  author={Deutch, Gilad and Gal, Rinon and Garibi, Daniel and Patashnik, Or and Cohen-Or, Daniel},
  booktitle={SIGGRAPH Asia 2024 Conference Papers},
  pages={1--12},
  year={2024}
}

@inproceedings{schwettmann2021toward,
  title={Toward a visual concept vocabulary for gan latent space},
  author={Schwettmann, Sarah and Hernandez, Evan and Bau, David and Klein, Samuel and Andreas, Jacob and Torralba, Antonio},
  booktitle={Proceedings of the IEEE/CVF International Conference on Computer Vision},
  pages={6804--6812},
  year={2021}
}

@inproceedings{goetschalckx2019ganalyze,
  title={Ganalyze: Toward visual definitions of cognitive image properties},
  author={Goetschalckx, Lore and Andonian, Alex and Oliva, Aude and Isola, Phillip},
  booktitle={Proceedings of the ieee/cvf international conference on computer vision},
  pages={5744--5753},
  year={2019}
}

@inproceedings{yuksel2021latentclr,
  title={Latentclr: A contrastive learning approach for unsupervised discovery of interpretable directions},
  author={Y{\"u}ksel, O{\u{g}}uz Kaan and Simsar, Enis and Er, Ezgi G{\"u}lperi and Yanardag, Pinar},
  booktitle={Proceedings of the IEEE/CVF international conference on computer vision},
  pages={14263--14272},
  year={2021}
}

@inproceedings{hu2022unsupervised,
  title={Unsupervised Discovery of Disentangled Interpretable Directions for Layer-Wise GAN},
  author={Hu, Haotian and Jiang, Bin and Zhou, Xinjiao and Huo, Xiaofei and Zhang, Bolin},
  booktitle={CCF Conference on Big Data},
  pages={21--39},
  year={2022},
  organization={Springer}
}

@inproceedings{yang2021discovering,
  title={Discovering interpretable latent space directions of gans beyond binary attributes},
  author={Yang, Huiting and Chai, Liangyu and Wen, Qiang and Zhao, Shuang and Sun, Zixun and He, Shengfeng},
  booktitle={Proceedings of the IEEE/CVF conference on computer vision and pattern recognition},
  pages={12177--12185},
  year={2021}
}

@article{harkonen2020ganspace,
  title={Ganspace: Discovering interpretable gan controls},
  author={H{\"a}rk{\"o}nen, Erik and Hertzmann, Aaron and Lehtinen, Jaakko and Paris, Sylvain},
  journal={Advances in neural information processing systems},
  volume={33},
  pages={9841--9850},
  year={2020}
}

@inproceedings{voynov2020unsupervised,
  title={Unsupervised discovery of interpretable directions in the gan latent space},
  author={Voynov, Andrey and Babenko, Artem},
  booktitle={International conference on machine learning},
  pages={9786--9796},
  year={2020},
  organization={PMLR}
}

@inproceedings{esser2024scaling,
  title={Scaling rectified flow transformers for high-resolution image synthesis},
  author={Esser, Patrick and Kulal, Sumith and Blattmann, Andreas and Entezari, Rahim and M{\"u}ller, Jonas and Saini, Harry and Levi, Yam and Lorenz, Dominik and Sauer, Axel and Boesel, Frederic and others},
  booktitle={Forty-first international conference on machine learning},
  year={2024}
}

@article{liu2022flow,
  title={Flow straight and fast: Learning to generate and transfer data with rectified flow},
  author={Liu, Xingchao and Gong, Chengyue and Liu, Qiang},
  journal={arXiv preprint arXiv:2209.03003},
  year={2022}
}

@article{lipman2022flow,
  title={Flow matching for generative modeling},
  author={Lipman, Yaron and Chen, Ricky TQ and Ben-Hamu, Heli and Nickel, Maximilian and Le, Matt},
  journal={arXiv preprint arXiv:2210.02747},
  year={2022}
}

@article{ho2020denoising,
  title={Denoising diffusion probabilistic models},
  author={Ho, Jonathan and Jain, Ajay and Abbeel, Pieter},
  journal={Advances in neural information processing systems},
  volume={33},
  pages={6840--6851},
  year={2020}
}

@inproceedings{orgad2023editing,
  title={Editing implicit assumptions in text-to-image diffusion models},
  author={Orgad, Hadas and Kawar, Bahjat and Belinkov, Yonatan},
  booktitle={Proceedings of the IEEE/CVF International Conference on Computer Vision},
  pages={7053--7061},
  year={2023}
}

@inproceedings{zhang2023sine,
  title={Sine: Single image editing with text-to-image diffusion models},
  author={Zhang, Zhixing and Han, Ligong and Ghosh, Arnab and Metaxas, Dimitris N and Ren, Jian},
  booktitle={Proceedings of the IEEE/CVF conference on computer vision and pattern recognition},
  pages={6027--6037},
  year={2023}
}

@inproceedings{kawar2023imagic,
  title={Imagic: Text-based real image editing with diffusion models},
  author={Kawar, Bahjat and Zada, Shiran and Lang, Oran and Tov, Omer and Chang, Huiwen and Dekel, Tali and Mosseri, Inbar and Irani, Michal},
  booktitle={Proceedings of the IEEE/CVF conference on computer vision and pattern recognition},
  pages={6007--6017},
  year={2023}
}

@inproceedings{dong2023prompt,
  title={Prompt tuning inversion for text-driven image editing using diffusion models},
  author={Dong, Wenkai and Xue, Song and Duan, Xiaoyue and Han, Shumin},
  booktitle={Proceedings of the IEEE/CVF international conference on computer vision},
  pages={7430--7440},
  year={2023}
}

@article{song2020denoising,
  title={Denoising diffusion implicit models},
  author={Song, Jiaming and Meng, Chenlin and Ermon, Stefano},
  journal={arXiv preprint arXiv:2010.02502},
  year={2020}
}

@inproceedings{jia2025designedit,
  title={DesignEdit: Unify Spatial-Aware Image Editing via Training-free Inpainting with a Multi-Layered Latent Diffusion Framework},
  author={Jia, Yueru and Cheng, Aosong and Yuan, Yuhui and Wang, Chuke and Li, Ji and Jia, Huizhu and Zhang, Shanghang},
  booktitle={Proceedings of the AAAI Conference on Artificial Intelligence},
  volume={39},
  number={4},
  pages={3958--3966},
  year={2025}
}

@inproceedings{wu2025latentps,
  title={LatentPS: Image Editing Using Latent Representations in Diffusion Models},
  author={Wu, Zilong and Murata, Hideki and Takahashi, Nayu and Wu, Qiyu and Tsuruoka, Yoshimasa},
  booktitle={Proceedings of the Winter Conference on Applications of Computer Vision},
  pages={167--176},
  year={2025}
}

@inproceedings{MacQueen1967,
  author = {MacQueen, J. B.},
  booktitle = {Proc. of the fifth Berkeley Symposium on Mathematical Statistics and Probability},
  editor = {Cam, L. M. Le and Neyman, J.},
  pages = {281-297},
  publisher = {University of California Press},
  title = {Some Methods for Classification and Analysis of Multivariate Observations},
  volume = 1,
  year = 1967
}

@article{nemhauser1978analysis,
  title={An analysis of approximations for maximizing submodular set functions—I},
  author={Nemhauser, George L and Wolsey, Laurence A and Fisher, Marshall L},
  journal={Mathematical programming},
  volume={14},
  number={1},
  pages={265--294},
  year={1978},
  publisher={Springer}
}

@article{sviridenko2004note,
  title={A note on maximizing a submodular set function subject to a knapsack constraint},
  author={Sviridenko, Maxim},
  journal={Operations Research Letters},
  volume={32},
  number={1},
  pages={41--43},
  year={2004},
  publisher={Elsevier}
}

@inproceedings{ye2024progressive,
  title={Progressive text-to-image diffusion with soft latent direction},
  author={Ye, YuTeng and Cai, Jiale and Zhou, Hang and Li, Guanwen and Zhang, Youjia and Song, Zikai and Gao, Chenxing and Yu, Junqing and Yang, Wei},
  booktitle={Proceedings of the AAAI Conference on Artificial Intelligence},
  volume={38},
  number={7},
  pages={6693--6701},
  year={2024}
}

@article{ibarrola2024measuring,
  title={Measuring diversity in co-creative image generation},
  author={Ibarrola, Francisco and Grace, Kazjon},
  journal={arXiv preprint arXiv:2403.13826},
  year={2024}
}

@inproceedings{hu2023tifa,
  title={Tifa: Accurate and interpretable text-to-image faithfulness evaluation with question answering},
  author={Hu, Yushi and Liu, Benlin and Kasai, Jungo and Wang, Yizhong and Ostendorf, Mari and Krishna, Ranjay and Smith, Noah A},
  booktitle={Proceedings of the IEEE/CVF International Conference on Computer Vision},
  pages={20406--20417},
  year={2023}
}

@inproceedings{radford2021learning,
  title={Learning transferable visual models from natural language supervision},
  author={Radford, Alec and Kim, Jong Wook and Hallacy, Chris and Ramesh, Aditya and Goh, Gabriel and Agarwal, Sandhini and Sastry, Girish and Askell, Amanda and Mishkin, Pamela and Clark, Jack and others},
  booktitle={International conference on machine learning},
  pages={8748--8763},
  year={2021},
  organization={PmLR}
}

@inproceedings{wu2023latent,
  title={A latent space of stochastic diffusion models for zero-shot image editing and guidance},
  author={Wu, Chen Henry and De la Torre, Fernando},
  booktitle={Proceedings of the IEEE/CVF International Conference on Computer Vision},
  pages={7378--7387},
  year={2023}
}

@inproceedings{xu2023versatile,
  title={Versatile diffusion: Text, images and variations all in one diffusion model},
  author={Xu, Xingqian and Wang, Zhangyang and Zhang, Gong and Wang, Kai and Shi, Humphrey},
  booktitle={Proceedings of the IEEE/CVF international conference on computer vision},
  pages={7754--7765},
  year={2023}
}

@article{saharia2022photorealistic,
  title={Photorealistic text-to-image diffusion models with deep language understanding},
  author={Saharia, Chitwan and Chan, William and Saxena, Saurabh and Li, Lala and Whang, Jay and Denton, Emily L and Ghasemipour, Kamyar and Gontijo Lopes, Raphael and Karagol Ayan, Burcu and Salimans, Tim and others},
  journal={Advances in neural information processing systems},
  volume={35},
  pages={36479--36494},
  year={2022}
}

@misc{Labs_2024, title={Announcing Black Forest Labs}, url={https://bfl.ai/blog/24-08-01-bfl}, author={Labs, Black Forest}, year={2024}, month=aug, note={Accessed on June 27, 2026}, language={en} }

@article{ramesh2022hierarchical,
  title={Hierarchical text-conditional image generation with clip latents},
  author={Ramesh, Aditya and Dhariwal, Prafulla and Nichol, Alex and Chu, Casey and Chen, Mark},
  journal={arXiv preprint arXiv:2204.06125},
  volume={1},
  number={2},
  pages={3},
  year={2022}
}

@inproceedings{rombach2022high,
  title={High-resolution image synthesis with latent diffusion models},
  author={Rombach, Robin and Blattmann, Andreas and Lorenz, Dominik and Esser, Patrick and Ommer, Bj{\"o}rn},
  booktitle={Proceedings of the IEEE/CVF conference on computer vision and pattern recognition},
  pages={10684--10695},
  year={2022}
}

@inproceedings{podell2024sdxl,
  title={Sdxl: Improving latent diffusion models for high-resolution image synthesis},
  author={Podell, Dustin and English, Zion and Lacey, Kyle and Blattmann, Andreas and Dockhorn, Tim and M{\"u}ller, Jonas and Penna, Joe and Rombach, Robin},
  booktitle={International Conference on Learning Representations},
  volume={2024},
  pages={1862--1874},
  year={2024}
}

@inproceedings{han2025enhancing,
  title={Enhancing creative generation on stable diffusion-based models},
  author={Han, Jiyeon and Kwon, Dahee and Lee, Gayoung and Kim, Junho and Choi, Jaesik},
  booktitle={Proceedings of the Computer Vision and Pattern Recognition Conference},
  pages={28609--28618},
  year={2025}
}

@article{liu2023pre,
  title={Pre-train, prompt, and predict: A systematic survey of prompting methods in natural language processing},
  author={Liu, Pengfei and Yuan, Weizhe and Fu, Jinlan and Jiang, Zhengbao and Hayashi, Hiroaki and Neubig, Graham},
  journal={ACM computing surveys},
  volume={55},
  number={9},
  pages={1--35},
  year={2023},
  publisher={ACM New York, NY}
}

@inproceedings{marvin2023prompt,
  title={Prompt engineering in large language models},
  author={Marvin, Ggaliwango and Hellen, Nakayiza and Jjingo, Daudi and Nakatumba-Nabende, Joyce},
  booktitle={International conference on data intelligence and cognitive informatics},
  pages={387--402},
  year={2023},
  organization={Springer}
}

@article{campbell2024understanding,
  title={Understanding the limits of vision language models through the lens of the binding problem},
  author={Campbell, Declan and Rane, Sunayana and Giallanza, Tyler and De Sabbata, Nicol{\`o} and Ghods, Kia and Joshi, Amogh and Ku, Alexander and Frankland, Steven M and Griffiths, Thomas L and Cohen, Jonathan D and others},
  journal={Advances in Neural Information Processing Systems},
  volume={37},
  pages={113436--113460},
  year={2024}
}

@article{izadi2026visual,
  title={Visual structures help visual reasoning: Addressing the binding problem in LVLMs},
  author={Izadi, Amirmohammad and Banayeeanzade, Mohammadali and Askari, Fatemeh and Rahimiakbar, Ali and Vahedi, Mohammad and Hasani, Hosein and Baghshah, Mahdieh},
  journal={Advances in Neural Information Processing Systems},
  volume={38},
  pages={162930--162968},
  year={2026}
}

@inproceedings{zhang2025critic,
  title={Critic-v: Vlm critics help catch vlm errors in multimodal reasoning},
  author={Zhang, Di and Lei, Jingdi and Li, Junxian and Wang, Xunzhi and Liu, Yujie and Yang, Zonglin and Li, Jiatong and Wang, Weida and Yang, Suorong and Wu, Jianbo and others},
  booktitle={Proceedings of the IEEE/CVF Conference on Computer Vision and Pattern Recognition},
  pages={9050--9061},
  year={2025}
}

@inproceedings{arora2023have,
  title={Have llms advanced enough? a challenging problem solving benchmark for large language models},
  author={Arora, Daman and Singh, Himanshu and others},
  booktitle={Proceedings of the 2023 Conference on Empirical Methods in Natural Language Processing},
  pages={7527--7543},
  year={2023}
}

@inproceedings{qu2023layoutllm,
  title={Layoutllm-t2i: Eliciting layout guidance from llm for text-to-image generation},
  author={Qu, Leigang and Wu, Shengqiong and Fei, Hao and Nie, Liqiang and Chua, Tat-Seng},
  booktitle={Proceedings of the 31st ACM International Conference on Multimedia},
  pages={643--654},
  year={2023}
}

@inproceedings{meral2024conform,
  title={Conform: Contrast is all you need for high-fidelity text-to-image diffusion models},
  author={Meral, Tuna Han Salih and Simsar, Enis and Tombari, Federico and Yanardag, Pinar},
  booktitle={Proceedings of the IEEE/CVF Conference on Computer Vision and Pattern Recognition},
  pages={9005--9014},
  year={2024}
}

@article{zhang2022divergan,
  title={DiverGAN: an efficient and effective single-stage framework for diverse text-to-image generation},
  author={Zhang, Zhenxing and Schomaker, Lambert},
  journal={Neurocomputing},
  volume={473},
  pages={182--198},
  year={2022},
  publisher={Elsevier}
}

@article{venkatesh2024ravel,
  title={RAVEL: Rare Concept Generation and Editing via Graph-driven Relational Guidance},
  author={Venkatesh, Kavana and Dalva, Yusuf and Lourentzou, Ismini and Yanardag, Pinar},
  journal={arXiv preprint arXiv:2412.09614},
  year={2024}
}

@inproceedings{liu2025medebench,
  title={MedEBench: Diagnosing Reliability in Text-Guided Medical Image Editing},
  author={Liu, Minghao and He, Zhitao and Fan, Zhiyuan and Wang, Qingyun and Fung, Yi R},
  booktitle={Findings of the Association for Computational Linguistics: EMNLP 2025},
  pages={767--791},
  year={2025}
}

@inproceedings{yang2026vibe,
  title={Vibe Spaces for Creatively Connecting and Expressing Visual Concepts},
  author={Yang, Huzheng and Xu, Katherine and Lu, Andrew and Grossberg, Michael D and Bai, Yutong and Shi, Jianbo},
  booktitle={Proceedings of the IEEE/CVF Conference on Computer Vision and Pattern Recognition},
  pages={21912--21921},
  year={2026}
}

@article{sivuk2025diverse,
  title={Diverse semantic image editing with style codes},
  author={Sivuk, Hakan and Dundar, Aysegul},
  journal={IEEE Transactions on Neural Networks and Learning Systems},
  year={2025},
  publisher={IEEE}
}

@inproceedings{xia2021tedigan,
  title={Tedigan: Text-guided diverse face image generation and manipulation},
  author={Xia, Weihao and Yang, Yujiu and Xue, Jing-Hao and Wu, Baoyuan},
  booktitle={Proceedings of the IEEE/CVF conference on computer vision and pattern recognition},
  pages={2256--2265},
  year={2021}
}

@inproceedings{choi2020stargan,
  title={Stargan v2: Diverse image synthesis for multiple domains},
  author={Choi, Yunjey and Uh, Youngjung and Yoo, Jaejun and Ha, Jung-Woo},
  booktitle={Proceedings of the IEEE/CVF conference on computer vision and pattern recognition},
  pages={8188--8197},
  year={2020}
}

@inproceedings{sonmezer2025loraverse,
  title={LoRAverse: A Submodular Framework to Retrieve Diverse Adapters for Diffusion Models},
  author={Sonmezer, Mert and Zheng, Matthew and Yanardag, Pinar},
  booktitle={Proceedings of the IEEE/CVF International Conference on Computer Vision},
  pages={17879--17888},
  year={2025}
}

@article{venkatesh2026crea,
  title={Crea: A collaborative multi-agent framework for creative image editing and generation},
  author={Venkatesh, Kavana and Dunlop, Connor and Yanardag, Pinar},
  journal={Advances in Neural Information Processing Systems},
  volume={38},
  pages={171332--171392},
  year={2026}
}

@inproceedings{chen2026failureatlas,
  title={Failureatlas: Mapping the failure landscape of t2i models via active exploration},
  author={Chen, Muxi and Zhang, Zhaohua and Zhao, Chenchen and Chen, Mingyang and Jiang, Wenyu and Jiang, Tianwen and Zhuo, Jianhuan and Tang, Yu and Xiao, Qiuyong and Zhang, Jihong and others},
  booktitle={Proceedings of the IEEE/CVF Conference on Computer Vision and Pattern Recognition},
  pages={40782--40791},
  year={2026}
}

@article{dunlop2026personalized,
  title={Personalized image editing in text-to-image diffusion models via collaborative direct preference optimization},
  author={Dunlop, Connor and Zheng, Matthew and Venkatesh, Kavana and Yanardag, Pinar},
  journal={Advances in Neural Information Processing Systems},
  volume={38},
  pages={167742--167771},
  year={2026}
}

@inproceedings{butt2026lumictrl,
  title={LumiCtrl: Learning Illuminant Prompts for Lighting Control in Personalized Text-to-Image Models},
  author={Butt, Muhammad Atif and Wang, Kai and Vazquez-Corral, Javier and Van De Weijer, Joost},
  booktitle={Proceedings of the IEEE/CVF Conference on Computer Vision and Pattern Recognition},
  pages={5925--5934},
  year={2026}
}

@inproceedings{li2026dit,
  title={Dit-vton: Diffusion transformer framework for unified multi-category virtual try-on and virtual try-all with integrated image editing},
  author={Li, Qi and Qiu, Shuwen and Koo, Kee Kiat and Han, Julien and Bouyarmane, Karim},
  booktitle={Proceedings of the IEEE/CVF Winter Conference on Applications of Computer Vision},
  pages={202--211},
  year={2026}
}

@inproceedings{choi2024improving,
  title={Improving diffusion models for authentic virtual try-on in the wild},
  author={Choi, Yisol and Kwak, Sangkyung and Lee, Kyungmin and Choi, Hyungwon and Shin, Jinwoo},
  booktitle={European Conference on Computer Vision},
  pages={206--235},
  year={2024},
  organization={Springer}
}

@inproceedings{parihar2024precisecontrol,
  title={Precisecontrol: Enhancing text-to-image diffusion models with fine-grained attribute control},
  author={Parihar, Rishubh and Sachidanand, VS and Mani, Sabariswaran and Karmali, Tejan and Venkatesh Babu, R},
  booktitle={European Conference on Computer Vision},
  pages={469--487},
  year={2024},
  organization={Springer}
}

@article{zhao2024ultraedit,
  title={Ultraedit: Instruction-based fine-grained image editing at scale},
  author={Zhao, Haozhe and Ma, Xiaojian and Chen, Liang and Si, Shuzheng and Wu, Rujie and An, Kaikai and Yu, Peiyu and Zhang, Minjia and Li, Qing and Chang, Baobao},
  journal={Advances in Neural Information Processing Systems},
  volume={37},
  pages={3058--3093},
  year={2024}
}

@inproceedings{baumann2025continuous,
  title={Continuous, subject-specific attribute control in t2i models by identifying semantic directions},
  author={Baumann, Stefan Andreas and Krause, Felix and Neumayr, Michael and Stracke, Nick and Sevi, Melvin and Hu, Vincent Tao and Ommer, Bj{\"o}rn},
  booktitle={Proceedings of the Computer Vision and Pattern Recognition Conference},
  pages={13231--13241},
  year={2025}
}

@inproceedings{chen2024artadapter,
  title={Artadapter: Text-to-image style transfer using multi-level style encoder and explicit adaptation},
  author={Chen, Dar-Yen and Tennent, Hamish and Hsu, Ching-Wen},
  booktitle={Proceedings of the IEEE/CVF conference on computer vision and pattern recognition},
  pages={8619--8628},
  year={2024}
}

@inproceedings{lin2011class,
  title={A class of submodular functions for document summarization},
  author={Lin, Hui and Bilmes, Jeff},
  booktitle={Proceedings of the 49th annual meeting of the association for computational linguistics: human language technologies},
  pages={510--520},
  year={2011}
}

@article{mazuz2025consistyle,
  title={ConsiStyle: Style Diversity in Training-Free Consistent T2I Generation},
  author={Mazuz, Yohai and Bruner, Janna and Wolf, Lior},
  journal={ACM Transactions on Graphics (TOG)},
  volume={44},
  number={6},
  pages={1--16},
  year={2025},
  publisher={ACM New York, NY, USA}
}

@inproceedings{Simsar_2023_WACV,
    author    = {Simsar, Enis and Kocasari, Umut and Er, Ezgi G\"ulperi and Yanardag, Pinar},
    title     = {Fantastic Style Channels and Where To Find Them: A Submodular Framework for Discovering Diverse Directions in GANs},
    booktitle = {Proceedings of the IEEE/CVF Winter Conference on Applications of Computer Vision (WACV)},
    month     = {January},
    year      = {2023},
    pages     = {4731-4740}
}

@inproceedings{han2025stylebooth,
  title={Stylebooth: Image style editing with multimodal instruction},
  author={Han, Zhen and Mao, Chaojie and Jiang, Zeyinzi and Pan, Yulin and Zhang, Jingfeng},
  booktitle={Proceedings of the IEEE/CVF International Conference on Computer Vision},
  pages={1947--1957},
  year={2025}
}

@inproceedings{lin2024evaluating,
  title={Evaluating text-to-visual generation with image-to-text generation},
  author={Lin, Zhiqiu and Pathak, Deepak and Li, Baiqi and Li, Jiayao and Xia, Xide and Neubig, Graham and Zhang, Pengchuan and Ramanan, Deva},
  booktitle={European Conference on Computer Vision},
  pages={366--384},
  year={2024},
  organization={Springer}
}

@inproceedings{hessel2021clipscore,
  title={Clipscore: A reference-free evaluation metric for image captioning},
  author={Hessel, Jack and Holtzman, Ari and Forbes, Maxwell and Le Bras, Ronan and Choi, Yejin},
  booktitle={Proceedings of the 2021 conference on empirical methods in natural language processing},
  pages={7514--7528},
  year={2021}
}

@inproceedings{gandikota2025sliderspace,
  title={Sliderspace: Decomposing the visual capabilities of diffusion models},
  author={Gandikota, Rohit and Wu, Zongze and Zhang, Richard and Bau, David and Shechtman, Eli and Kolkin, Nick},
  booktitle={Proceedings of the IEEE/CVF International Conference on Computer Vision},
  pages={15994--16003},
  year={2025}
}

@inproceedings{dalva2024noiseclr,
  title={Noiseclr: A contrastive learning approach for unsupervised discovery of interpretable directions in diffusion models},
  author={Dalva, Yusuf and Yanardag, Pinar},
  booktitle={Proceedings of the IEEE/CVF conference on computer vision and pattern recognition},
  pages={24209--24218},
  year={2024}
}

@article{zhang2023text,
  title={Text-to-image diffusion models in generative ai: A survey},
  author={Zhang, Chenshuang and Zhang, Chaoning and Zhang, Mengchun and Kweon, In So and Kim, Junmo},
  journal={arXiv preprint arXiv:2303.07909},
  year={2023}
}

@article{lin2024ctrl,
  title={Ctrl-x: Controlling structure and appearance for text-to-image generation without guidance},
  author={Lin, Kuan Heng and Mo, Sicheng and Klingher, Ben and Mu, Fangzhou and Zhou, Bolei},
  journal={Advances in Neural Information Processing Systems},
  volume={37},
  pages={128911--128939},
  year={2024}
}

@article{li2019controllable,
  title={Controllable text-to-image generation},
  author={Li, Bowen and Qi, Xiaojuan and Lukasiewicz, Thomas and Torr, Philip},
  journal={Advances in neural information processing systems},
  volume={32},
  year={2019}
}

@article{singh2025openai,
  title={Openai gpt-5 system card},
  author={Singh, Aaditya and Fry, Adam and Perelman, Adam and Tart, Adam and Ganesh, Adi and El-Kishky, Ahmed and McLaughlin, Aidan and Low, Aiden and Ostrow, AJ and Ananthram, Akhila and others},
  journal={arXiv preprint arXiv:2601.03267},
  year={2025}
}

@inproceedings{szegedy2016rethinking,
  title={Rethinking the inception architecture for computer vision},
  author={Szegedy, Christian and Vanhoucke, Vincent and Ioffe, Sergey and Shlens, Jon and Wojna, Zbigniew},
  booktitle={Proceedings of the IEEE conference on computer vision and pattern recognition},
  pages={2818--2826},
  year={2016}
}

@inproceedings{haas2024discovering,
  title={Discovering interpretable directions in the semantic latent space of diffusion models},
  author={Haas, Ren{\'e} and Huberman-Spiegelglas, Inbar and Mulayoff, Rotem and Gra{\ss}hof, Stella and Brandt, Sami S and Michaeli, Tomer},
  booktitle={2024 IEEE 18th International Conference on Automatic Face and Gesture Recognition (FG)},
  pages={1--9},
  year={2024},
  organization={IEEE}
}

@article{zhang2023unsupervised,
  title={Unsupervised discovery of interpretable directions in h-space of pre-trained diffusion models},
  author={Zhang, Zijian and Liu, Luping and Lin, Zhijie and Zhu, Yichen and Zhao, Zhou},
  journal={arXiv preprint arXiv:2310.09912},
  year={2023}
}

\clearpage
\appendix
\section*{Supplementary Material}
\addcontentsline{toc}{section}{Supplementary Material}

In this appendix, we present qualitative and quantitative experiments that demonstrate the generalizability of the top semantics. We also provide the prompts used to generate the lists of potential semantics and more details regarding the time comparisons and hyperparameters.

\section{Generalizability of Semantics}
\begin{figure}[h] 
  \centering
  \includegraphics[width=\linewidth]{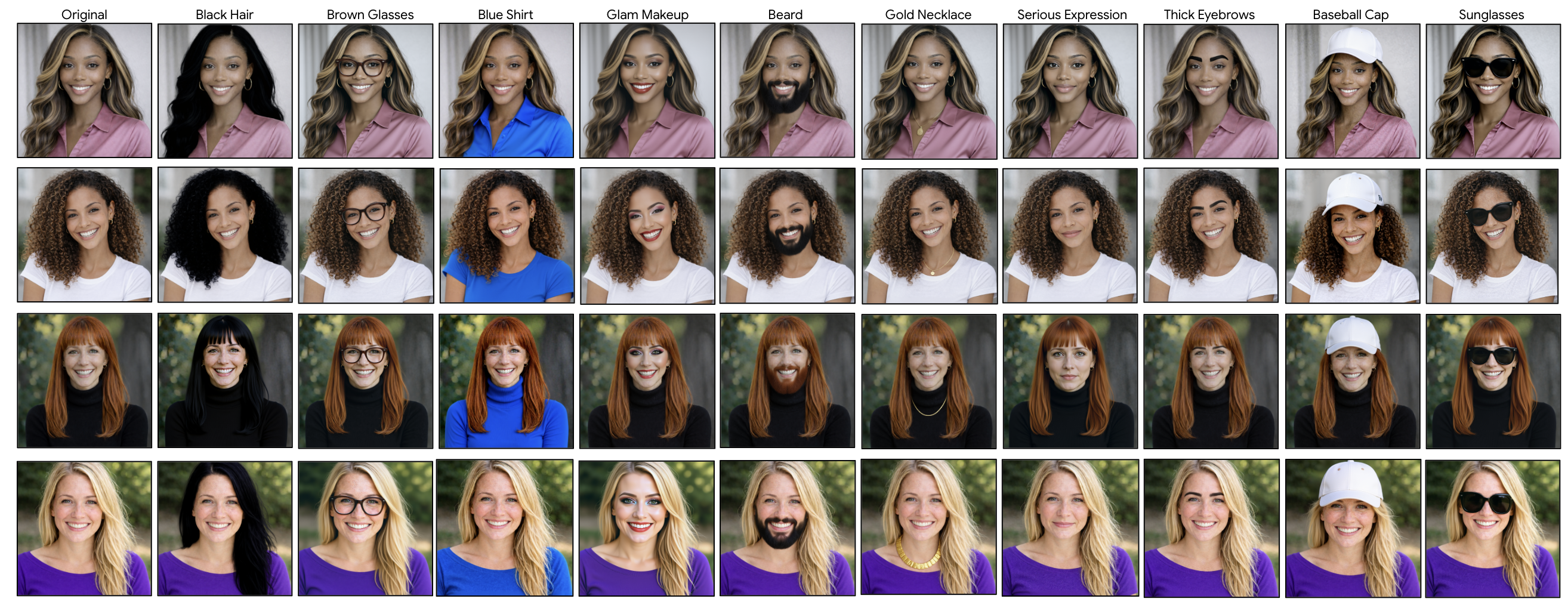}
  \caption{\textbf{Generalizability of Woman Semantics.} This image shows examples of the top 10 woman semantics discovered by our method using FLUX.1 Kontext and how they can be applied to edit different images.}
  \label{fig:womangen}
\end{figure}

Figure \ref{fig:womangen} shows additional generalizability qualitative experiments in the face domain (women) using FLUX.1 Kontext \cite{labs2025flux}. Table~\ref{tab:generalizability} further validates this observation through quantitative interpretability metrics, including CLIP-T \cite{radford2021learning}, TIFA \cite{hu2023tifa}, and VQAScore \cite{lin2024evaluating} (higher values indicate stronger alignment). For these experiments, we used 2,500 images per base model (e.g., FLUX.1 Kontext [dev] and Stable Diffusion~1.5), resulting in a total of 5,000 images. Specifically, for each model, we used the top 25 semantic directions from 100 images, yielding 2,500 edited samples per model. As shown in Table~\ref{tab:generalizability}, RankT2I achieves high CLIP-T, TIFA, and VQAScores, indicating strong interpretability and text–image alignment. These improvements suggest that the semantic edits produced by RankT2I reliably match the intended prompts across diverse images, demonstrating the generalizability of the discovered semantics.

\section{Prompts}
We include the prompt template used to generate a list of candidate semantics as part of Step 1 of our method in Table \ref{tab:prompt_template}. Table \ref{tab:prompt_ruins} provides an example prompt for the \textit{ancient ruins} domain, while Table~\ref{tab:ruins} presents a subset of the resulting semantics generated from this prompt. To encourage high-quality and semantically diverse prompts, we provide GPT-5 \cite{singh2025openai} with example semantics for each broad attribute category (e.g., \textit{sunrise}, \textit{sunset}, and \textit{morning} for the category \textit{time of day}). This structured prompting strategy enables the generation of diverse yet coherent editing instructions, which serve as candidate semantics for downstream ranking and editing.

\section{Time Comparisons and Image Edit Details}
The total time needed to discover 100 directions using RankT2I, NoiseCLR, and SliderSpace depends on various factors. When performing the time comparisons as seen in the main paper, we used 2,500 images as the number of images in the domain training dataset in NoiseCLR. For SliderSpace, we used 1,000 images to train the sliders. As for RankT2I, we used 5 unedited images as the base and generated 100 edits for each of the original images for a total of 500 images. Increasing or decreasing the number of images may affect the time needed to find the directions, but the key point we're trying to address is that RankT2I can find high-quality interpretable directions using fewer images and in less time. 

Regarding the experimental details for getting the qualitative images in Figures 4 and 5 in the main paper, we used the first-K tokens in NoiseCLR and first-K directions in SliderSpace to generate the images. The directions we used for the images in the art and face domain for NoiseCLR were from the official project repository, while we trained the directions for the landscape sliders using 2,500 images from the Places365 dataset \cite{zhou2017places}. Additionally, we used edit strengths between 2-10 and edit timesteps between 10-30 and 35-50. As for SliderSpace, we trained the directions for the art, face, and landscape domains using 1000 generated images, while the other directions were from the official project repository. We used edit scales between 0.5-5 in SliderSpace. These hyperparameters were selected for each domain and fine-tuned to produce the edits while preserving the overall layout of the original image and minimizing visual distortions. As for the images used in RankT2I, we used edit scales between 5-25 for the images generated by Ledits++ \cite{brack2024ledits++}, 4-9 for the images generated by InstructPix2Pix \cite{brooks2023instructpix2pix}, and 1-6 for the images generated by FLUX.1 Kontext \cite{labs2025flux}.

\begin{table}[t]
\centering
\footnotesize
\setlength{\tabcolsep}{5pt}
\renewcommand{\arraystretch}{1.05}
\begin{tabular}{lccc}
\toprule
\textbf{Model}
& $\uparrow$\textbf{CLIP-T}
& $\uparrow$\textbf{TIFA}
& $\uparrow$\textbf{VQA} \\
\midrule
SD   & 0.282 & 0.724 & 0.897 \\
FLUX & 0.276 & 0.799 & 0.912 \\
\bottomrule
\end{tabular}
\caption{\textbf{Generalization of Semantics across Base Models.}
Higher image--text alignment and visual question answering scores indicate that RankT2I's semantic edits consistently match the intended prompts across different images, demonstrating the generalizability of the discovered semantics.}
\label{tab:generalizability}
\end{table}

\begin{table}[t]
\centering
\small
\begin{tabular}{p{\linewidth}}
\toprule
\textbf{Prompt Template} \\
\midrule

{\ttfamily
Generate a Python string array containing distinct prompts for editing an image of <SUBJECT>. The prompts should span a wide range of attribute categories to produce diverse variations of the scene.
}

\medskip

\textbf{Example Attribute Structure:}

\textbf{Broad Category 1:} attribute 1, attribute 2, attribute 3, \ldots\\
\textbf{Broad Category 2:} attribute 1, attribute 2, attribute 3, \ldots\\
\textbf{Broad Category 3:} attribute 1, attribute 2, attribute 3, \ldots\\
\textbf{Broad Category 4:} attribute 1, attribute 2, attribute 3, \ldots\\
\textbf{etc.}

\\
\bottomrule
\end{tabular}
\caption{\textbf{Prompt Template.} The template is used to generate diverse editing prompts spanning multiple semantic attribute categories.}
\label{tab:prompt_template}
\end{table}

\begin{table}[t]
\centering
\small
\begin{tabular}{p{\linewidth}}
\toprule
\textbf{Prompt Example (Ancient Ruins)} \\
\midrule

{\ttfamily
Generate a Python string array containing distinct prompts for editing an image of ancient ruins. The prompts should span a wide range of attribute categories (e.g., time of day, weather, ruins condition, ruins design, style, etc.).
}

\medskip

\textbf{Example Attribute Categories:}

\textbf{Time of day:} sunrise, sunset, night, etc.\\
\textbf{Weather:} bright and sunny, overcast, rainy, etc.\\
\textbf{Ruins design:} faded murals, domed structures, etc.\\
\textbf{Style:} sepia-toned, monochrome, etc.\\
\textbf{etc.}

\\
\bottomrule
\end{tabular}
\caption{\textbf{Example Prompt Used to Retrieve a List of Potential Attributes.} This table shows an example of the prompt used to retrieve a list of potential semantics for the \textit{ancient ruins} domain.}
\label{tab:prompt_ruins}
\end{table}

\begin{table}[t]
\centering
\footnotesize
\begin{tabular}{p{\linewidth}}
\toprule
\textbf{Potential List of Semantics (Ancient Ruins)} \\
\midrule

\begin{verbatim}
"Make the ruins at sunrise",\\
"Make the ruins at sunset",\\
"Make the ruins in the early morning",\\
"Make the ruins at noon",\\
"Make the ruins in the afternoon",\\
"Make the ruins in the evening",\\
"Make the ruins at dusk",\\
"Make the ruins at night",\\
"Make the ruins under a full moon",\\
"Make the ruins bright and sunny",\\
"Make the ruins overcast",\\
"Make the ruins rainy",\\
"Make the ruins misty",\\
"Make the ruins dust-filled",\\
"Make the ruins with broken columns",\\
"Make the ruins with fallen pillars",\\
"Make the ruins with cracked stone walls",\\
"Make the ruins with collapsed arches",\\
"Make the ruins with scattered stone rubble",\\
"Make the ruins partially buried in sand",\\
"Make the ruins submerged in shallow water",\\
"Make the ruins heavily eroded",\\
"Make the ruins structurally intact",\\
"Make the ruins crumbling",\\
"Make the ruins with ancient runes",\\
"Make the ruins with carved stone patterns",\\
"Make the ruins with faded murals",\\
"Make the ruins with engraved symbols",\\
"Make the ruins with monumental statues",\\
\ldots\\
] \end{verbatim}

\\
\bottomrule
\end{tabular}
\caption{\textbf{Example List of Potential Semantics.} Examples of prompts used to edit images of \textit{ancient ruins}, spanning diverse attribute categories including time of day, weather, structural condition, and decorative elements, are shown above.}
\label{tab:ruins}
\end{table}

\end{document}